\documentclass[10pt,twocolumn,letterpaper]{article}

\usepackage{multirow}

\usepackage{iccv}              

\definecolor{iccvblue}{rgb}{0.21,0.49,0.74}
\usepackage[pagebackref,breaklinks,colorlinks,allcolors=iccvblue]{hyperref}

\def\paperID{6330} 
\def\confName{ICCV}
\def\confYear{2025}

\title{LEG-SLAM: Language-Enhanced Gaussian Splatting for Real-Time SLAM}

\author{
Roman Titkov\textsuperscript{1,\dag},
Egor Zubkov\textsuperscript{1},
Dmitry Yudin\textsuperscript{1,2},
Jaafar Mahmoud\textsuperscript{3},\\
Malik Mohrat\textsuperscript{3},
Gennady Sidorov\textsuperscript{3}
}

\begin{document}
\maketitle

\begingroup
\renewcommand\thefootnote{}\footnotetext{
\\
\noindent
\hspace*{1em}%
\textsuperscript{1}Center for Cognitive Modeling, Moscow Institute of Physics and Technology, Russia\\
\hspace*{1em}%
\textsuperscript{2}AIRI, Moscow, Russia\\
\hspace*{1em}%
\textsuperscript{3}Sberbank of Russia, Robotics Center, Moscow, Russia\\
\hspace*{1em}%
\textsuperscript{\dag}Corresponding author: titkov.re@phystech.edu
}
\endgroup
\begin{abstract}
Modern Gaussian Splatting methods have proven highly effective for real-time photorealistic rendering of 3D scenes. However, integrating semantic information into this representation remains a significant challenge, especially in maintaining real-time performance for SLAM (Simultaneous Localization and Mapping) applications. In this work, we introduce LEG-SLAM — a novel approach that fuses an optimized Gaussian Splatting implementation with visual-language feature extraction using DINOv2 followed by
learnable feature compressor based on Principal component analysis,
while enabling an online dense SLAM. Our method simultaneously generates high-quality photorealistic images and semantically labeled scene maps, achieving real-time scene reconstruction with more than 10 fps on the Replica dataset and 18 fps on ScanNet.
Experimental results show that our approach significantly outperforms state-of-the-art methods in reconstruction speed while achieving competitive rendering quality. The proposed system eliminates the need for prior data preparation such as camera's ego motion or pre-computed static semantic maps. 
With its potential applications in autonomous robotics, augmented reality, and other interactive domains, LEG-SLAM represents a significant step forward in real-time semantic 3D Gaussian-based SLAM.
Project page: \url{https://titrom025.github.io/LEG-SLAM/}

\end{abstract}    

\section{Introduction}
\label{sec:intro}

In recent years, 3D Gaussian Splatting has become a popular method for representing and rendering novel views in 3D scenes, especially in applications that demand high performance and real-time processing. However, traditional approaches, which are primarily focused on photorealistic rendering, often fail short to simultaneously deliver high-quality semantic information—a critical requirement for applications in autonomous driving, augmented reality, and robotics.

By real-time, we mean a delay in the operation of the algorithm that does not exceed the period of receipt of the input sensory data. For example, for indoor robots, the frequency of image reception is typically 10 frames per second.

\begin{figure}[t]
    \centering
    \includegraphics[width=0.47\textwidth]{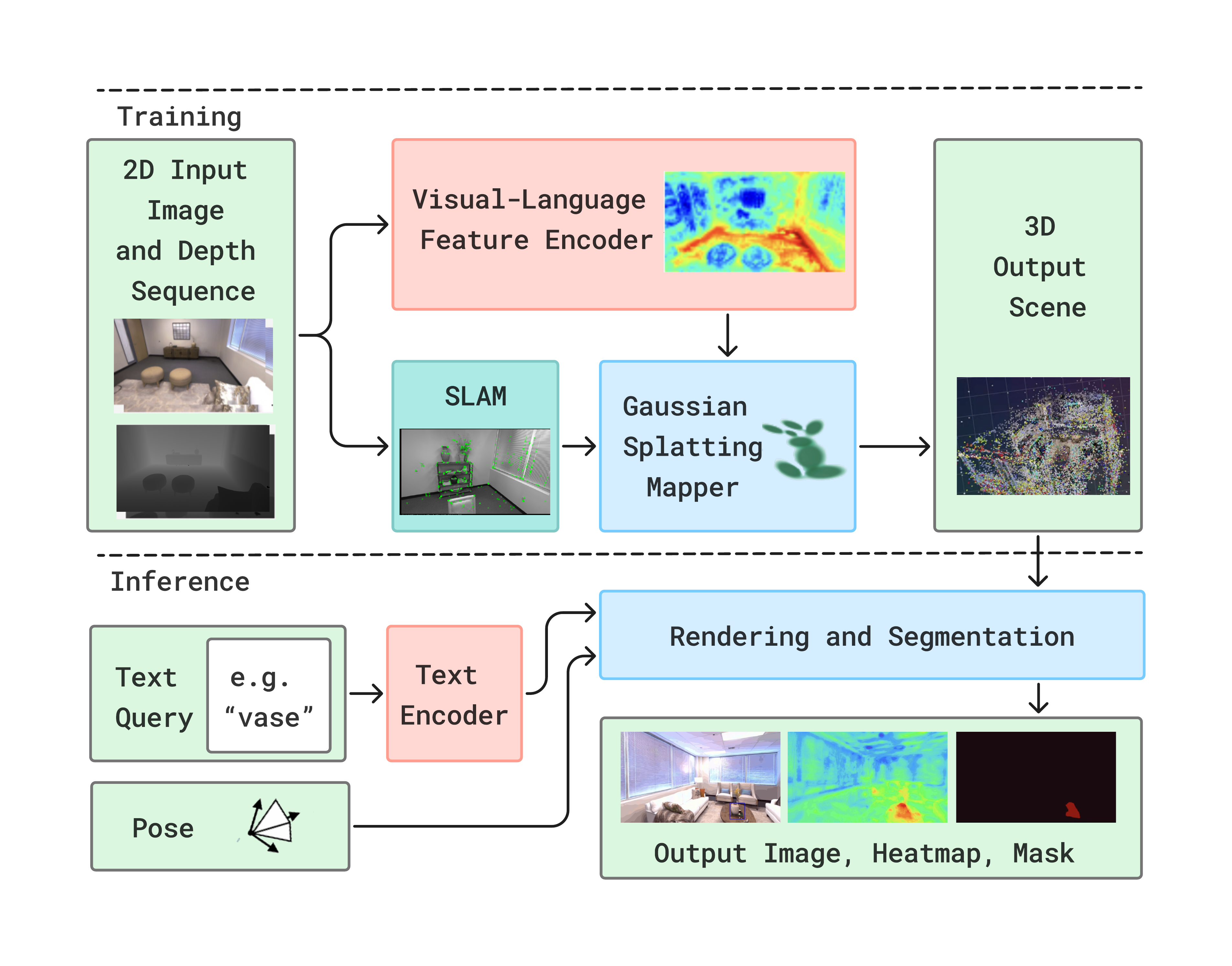}
    \caption{
    Simplified diagram of the proposed LEG-SLAM approach implementing real-time SLAM with language-enhanced Gaussian Splatting
    }
    \label{fig:GA}
\end{figure}

Following these advantages of the 3D representation, recent works  such as LangSplat~\cite{qin_langsplat_2024}, demonstrate the possibility of integrating language embeddings to support open-vocabulary queries. Despite its high semantic quality, LangSplat suffers from low performance in scene construction, limiting its real-time applicability. A similar approach, presented in~\cite{li_sgs-slam_2025}, integrates semantic features into 3D Gaussian Splatting for accurate mapping and tracking; however, it operates in a closed-vocabulary mode and requires a pre-prepared semantic map, which reduces its adaptability for mapping unseen environments. 

In this work, we present the first real-time SLAM (Simultaneous Localization and Mapping) method with visual-language features named \textbf{LEG-SLAM} (Language-Enhanced Gaussian Splatting SLAM). Our approach simultaneously renders photorealistic RGB images and creates accurate semantic masks, while executing a dense SLAM system in real-time. To achieve this, we have developed an efficient implementation of Gaussian Splatting, integrated with a semantic extraction module based on DINOv2 \cite{oquab_dinov2_2024} features from RGB images and Principal component analysis for feature compression. This approach significantly reduces processing time, saves memory and provides acceptable segmentation accuracy, making the system suitable for real-time operation.

One of the main advantages of our proposed method is the ability to construct a semantic representation of a 3D scene in real-time, without the need for prior data preparation. 
Unlike existing solutions that require static semantic maps or extensive pre-processing, our approach processes incoming RGB-D images on the fly, automatically extracting semantic features and building the 3D scene map.


\textbf{Our main contributions:}
\begin{itemize}
\item We developed an efficient Gaussian Splatting algorithm with original visual-language feature encoding based on DINOv2  and Principal Component Analysis (PCA).
\item We implemented the first real-time SLAM with language features in an unified manner, enabling open-vocabulary segmentation without the need for a pre-prepared static semantic map. It allowed us to achieve up to 30 times faster performance than that of the analogues.
\item We achieved real-time map construction for input images frame rate of 10 fps and competitive results in semantic segmentation quality on popular ScanNet and Replica datasets.
\end{itemize}
\section{Related work}
\label{sec:related_work}
\subsection{3D Scene Representation Techniques}
\label{subsec:3d_scene_representation}
Historically, 3D scenes have been represented using point clouds, meshes, and voxels. Point clouds are collections of points with XYZ coordinates and optional attributes like color. Meshes use polygons, usually triangles, to model object surfaces. Voxels divide 3D space into cubic elements analogous to 2D pixels.

Recently, neural implicit representations such as Neural Radiance Fields (NeRF) \cite{mildenhall_nerf_2020} have gained popularity. NeRF models a continuous 3D radiance field using neural networks, enabling highly photorealistic scene reconstructions. However, it is computationally intensive. Extensions like Semantic-NeRF \cite{zhi_-place_2021} and EditNeRF \cite{liu_editing_2021} add semantic labels and editing capabilities but remain impractical for real-time use.

On the other hand, 3D Gaussian Splatting (3DGS) \cite{kerbl_3d_2023} is a promising new representation that uses 3D Gaussians. It achieves photorealistic 1080p rendering at 60 FPS by leveraging point-based alpha blending and a differentiable tiled rasterizer. The explicit Gaussian parameterization also facilitates efficient scene editing. Dynamic 3D Gaussians \cite{luiten_dynamic_2023} extends 3DGS to handle dynamic scenes. Other works~\cite{yi_gaussiandreamer_2024, jiang_brightdreamer_2024} generate 3DGS scenes from text and images.

\subsection{Semantic 3D Scene Understanding} Large annotated datasets and multimodal models like CLIP \cite{radford_learning_2021} have advanced open-vocabulary semantic segmentation methods such as OpenSeg \cite{ghiasi_scaling_2022} and LSeg \cite{li_language-driven_2022}. However, most operate on single images without ensuring multi-view consistency, limiting their utility for 3D scene analysis.

NeRF-based approaches like Semantic-NeRF \cite{zhi_-place_2021} and Panoptic Lifting \cite{siddiqui_panoptic_2022} incorporate semantic embeddings for enhanced 3D scene understanding but suffer from high computational complexity. Recent works \cite{guo_semantic_2024, li_sgs-slam_2025} extend 3DGS with semantic embeddings for open-vocabulary segmentation and reconstruction, but remain computationally demanding, which prevents the real-time use.

\subsection{Neural Representations in SLAM} Neural representations have improved the reconstruction quality, speed, and semantic understanding of SLAM systems. SplaTAM~\cite{keetha_splatam_nodate} uses silhouette-guided optimization for better reconstruction. GS-SLAM~\cite{yan_gs-slam_2024} and GS-ICP~\cite{ha_rgbd_2024} integrate adaptive Gaussian expansion and G-ICP pose tracking for accuracy and scalability.

Hybrid neural-coordinate representations address efficiency issues. Co-SLAM~\cite{wang_co-slam_2023} combines multi-resolution hash grids and one-blob encoding for real-time global bundle adjustment without keyframes. E-SLAM~\cite{johari_eslam_2023} decodes multi-scale axis-aligned feature planes into TSDF and RGB, achieving state-of-the-art accuracy and speed. Photo-SLAM~\cite{huang_photo-slam_2024} connects explicit geometry and implicit photometric features via hyper primitives and Gaussian-Pyramid training, surpassing existing methods in quality and efficiency.

For semantic mapping, DNS SLAM~\cite{li_dns_2023} leverages 2D semantic priors and multi-view constraints. SNI-SLAM~\cite{zhu_sni-slam_2024} refines scene representations through feature collaboration and one-way correlation decoding. 3D Gaussian frameworks like SGS-SLAM~\cite{li_sgs-slam_2025} incorporate semantic colored labels but may miss higher-level semantics. SemGauss-SLAM~\cite{zhu_semgauss-slam_2024} and NEDS-SLAM~\cite{ji_neds-slam_2024} embed low-dimensional semantic features into 3D Gaussians but rely on 2D segmentation, limiting open-set robustness.

In summary, neural representations have revolutionized 3D scene understanding and SLAM. Gaussian-based techniques show great promise for efficient, high-quality, and semantically-aware scene representation. However, further research is needed to develop real-time open-vocabulary understanding while preserving multi-view consistency and scalability.

\begin{figure*}[t]
    \centering
    \includegraphics[width=0.99\textwidth]{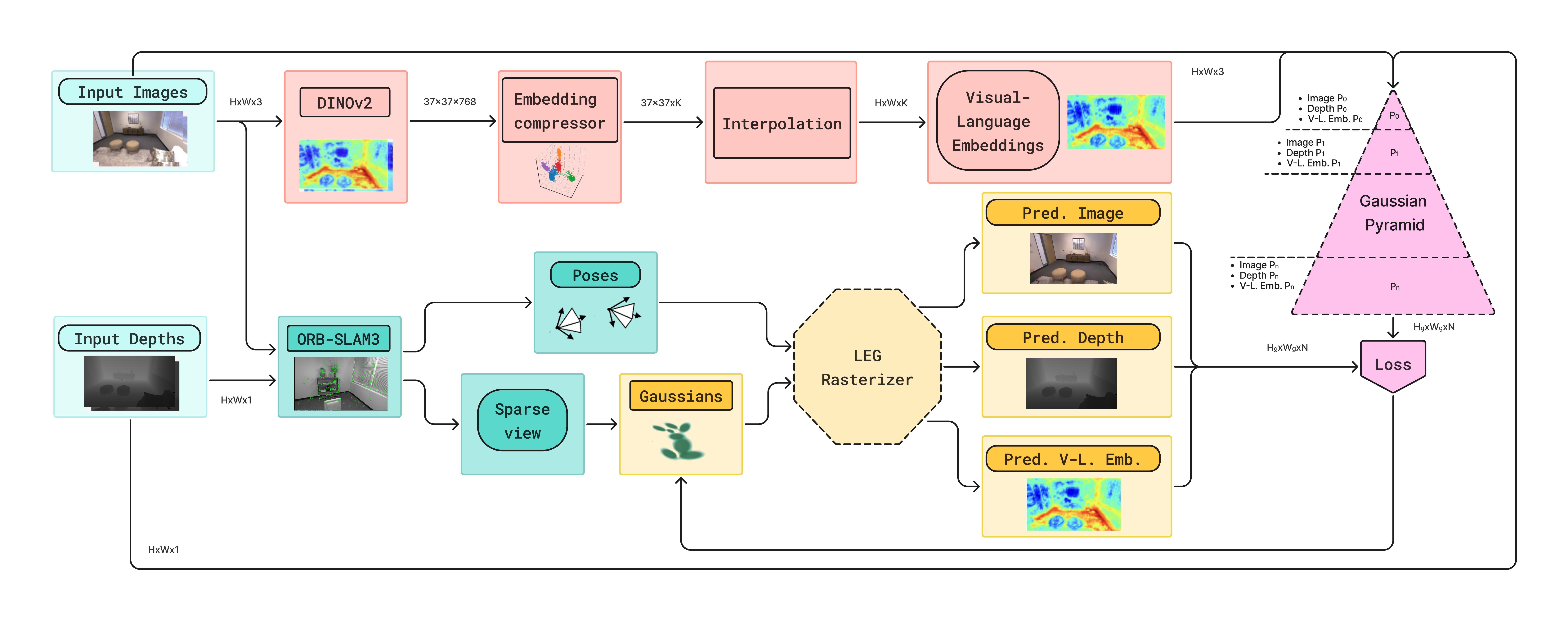}
    \caption{
    A detailed scheme of the developed LEG-SLAM method. Its distinctive features include a fast approach to obtaining visual-linguistic features, a computationally efficient SLAM implementation, and a learning-based approach to a language-enhanced rasterizer for Gaussian Splatting. 
    }
    \label{fig:method_overview}
\end{figure*}


\section{Methodology}
\label{sec:methodology}
\subsection{Proposed approach}

Unlike traditional approaches, where RGB image rendering and semantic segmentation are performed separately, our method integrates these processes into a unified pipeline, ensuring consistency between the visual and semantic representation of the 3D scene.
Figure~\ref{fig:method_overview} provides an overview of the proposed method. 
At the core of our approach lies the Gaussian Splatting method, which is used for efficient 3D scene representation and rendering. To extract visual features, we employ the \mbox{DINOv2} model~\cite{oquab_dinov2_2024}, which generates a compact embedding map characterizing local scene features. These embeddings are then compressed using an autoencoder, significantly reducing data size and accelerating subsequent processing.

The integration of text queries into the semantic segmentation process is performed using the Talk2DINO module~\cite{barsellotti_talking_2024}, which converts text embeddings from CLIP~\cite{radford_learning_2021}  space into the DINOv2~\cite{oquab_dinov2_2024} embedding space. This process is applied to an already reconstructed 3D scene, enabling open-vocabulary queries, where the semantic features extracted from the image are matched with textual descriptions in real time.

By combining all these modules into a single rasterization-based pipeline, our method enables the simultaneous output of high-quality RGB images and semantic masks, a crucial feature for real-time SLAM systems where minimizing latency is essential. The computational efficiency of our approach is detailed in Table~\ref{tab:performance_comparison}, which presents the average processing time per frame across different resolutions. The results highlight the scalability of our method, demonstrating that even at higher resolutions, the total processing time remains within real-time constraints.

\begin{table}[ht]
\centering
\small
\begin{tabular}{l|c|c|c}
\hline
\multicolumn{4}{c}{Processing Time Per Frame (ms)} \\
\hline
\textbf{Processing Stage} & \textbf{640×480} & \textbf{986×728}  & \textbf{1200×680} \\
\hline
Image preprocessing & 8.72 & 12.36 & 14.57 \\
Feature extraction & 31.91 & 40.38 & 48.33 \\
Feature compression & 1.57 & 2.14 & 2.36 \\
Tracking + Mapping & 14.82 & 26.85 & 29.95 \\
\hline
Average processing time & 57.02 & 78.65 & 95.21 \\
\hline
\end{tabular}%
\caption{Average processing time for each pipeline stage per frame at different resolutions.}
\label{tab:performance_comparison}
\end{table}

The implementation details of each component are provided in the following sections. Section~\ref{sec:visual_features_dinov2} describes the process of obtaining visual features using DINOv2~\cite{oquab_dinov2_2024}, while Section~\ref{sec:embedding_compressor} explains the embedding compression step using the autoencoder. The distillation process of semantic information into the scene representation is detailed in Section~\ref{sec:semantic_integration}, and the procedure for generating semantic queries using Talk2DINO~\cite{barsellotti_talking_2024}  is covered in Section~\ref{sec:segmentation_requests}.

\subsection{Visual features generation}
\label{sec:visual_features_dinov2}

To extract visual features, our system utilizes a pretrained DINOv2~\cite{oquab_dinov2_2024} model, which takes an input image of size \(518 \times 518\) pixels and produces an embedding map of size \(37 \times 37 \times 768\). In this map, each of the \(37 \times 37\) cells corresponds to a local scene fragment, while the 768-dimensional vector represents a high-level description of its visual characteristics.

For further processing, the embedding map is passed to the next system module without resizing. This approach allows working with a compact scene representation, reducing computational costs. To decrease the dimensionality of embeddings, an encoder compresses them from 768 to K dimensions. This significantly reduces computational complexity during subsequent processing while maintaining most of the essential information. The details of this compression are discussed in the next section.

Using DINOv2 enables the formation of an informative semantic representation, which is then distilled into the Gaussian Splatting pipeline. This ensures high segmentation quality while maintaining real-time processing capabilities.

\subsection{Embedding Compressor}
\label{sec:embedding_compressor}

To optimize the processing of embeddings generated by the DINOv2~\cite{oquab_dinov2_2024} model, our system employs a Principal Component Analysis (PCA)-based compressor. This module reduces the dimensionality of the embedding map from \(37 \times 37 \times 768\) to \(37 \times 37 \times K\), significantly lowering computational costs while preserving essential semantic information.

The PCA model was trained using 1,000 ImageNet text classes. For each class, a text embedding was obtained via CLIP~\cite{radford_learning_2021}, projected into the DINOv2~\cite{oquab_dinov2_2024} latent space using Talk2DINO~\cite{barsellotti_talking_2024}, and augmented with noise while maintaining a cosine similarity of at least 0.95 with the original. This strategy allows PCA to efficiently compress embeddings beyond ImageNet classes by learning a general feature projection rather than specific class representations.

After compression, the embedding map is upsampled to the original image resolution using bilinear interpolation, ensuring spatial consistency with the input image. The resulting \(H \times W \times K\) representation is then used in the Gaussian Splatting pipeline for joint rendering of RGB images and semantic maps. This approach enables real-time performance while preserving segmentation accuracy.

\subsection{Integration of Semantic Features and Gaussian Splatting}
\label{sec:semantic_integration}


In our approach, semantic features are incorporated to the Gaussian splatting pipeline for simultaneous rendering of RGB images and semantic masks. This allows to effectively combine color and semantic information in a scene, ensuring consistency of representations and high performance.


Semantic vector representing local features from the semantic embedding map is added to the initial parameters of each Gaussian. This semantic vector is optimized simultaneously with gaussian parameters (such as position, scale, orientation and color), which allows to obtain an accurate match between visual and semantic data. 

The optimization of semantic features vector includes the following steps:
\begin{enumerate}
    \item \textbf{Initialization of semantic vectors:} During the initial initialization of the gaussian, the semantic vector \(H\times W\times K\) is assigned zero values. If the gaussian is obtained by splitting of an existing gaussian, then the semantic features are copied.
    \item \textbf{Parameter optimization:} During the training process, the gaussian parameters are optimized by minimizing the error between the rendered and the reference 2d representations. Simultaneous training on a single gaussian cloud ensures consistency of visual and semantic data.
    \item \textbf{Simultaneous rendering of RGB and semantic masks:} The resulting gaussian parameters are used to simultaneously form a photorealistic image of a scene and corresponding semantic masks, where each gaussian contributes to both representations.
\end{enumerate}


Using a single pipeline for visual and semantic data can significantly reduce computational costs, while ensuring high accuracy and speed of reconstruction.


The loss function includes three terms. The first of them $\mathcal{L}_{color}$ is the standard loss function for color features from the original 3DGS method. The second term $\mathcal{L}_{cos\_sim}$ is responsible for calculating the cosine similarity between the rendered semantic feature map and the one obtained from DINOv2. The third term $\mathcal{L}_{1}$ optimizes the depth value. The inputs of the function are the rendered image $I_{pred}$, the language feature map $L_{pred}$ and the depth $D_{pred}$, as well as the gt image $I_{gt}$, the depth $D_{gt}$ and the compressed DINOv2 features  $L_{gt}$.
\begin{multline}
\mathcal{L}(I_{pred},\,L_{pred},\,D_{pred};\,I_{gt},\,L_{gt},\,D_{gt}) = \\ = \underbrace{(1- \lambda)\mathcal{L}_1(I_{pred},\,I_{gt})
+ \lambda\mathcal{L}_{ssim}(I_{pred},\,I_{gt})}_{\mathcal{L}_{color}}+ \\ + \mathcal{L}_{cos\_sim}(L_{pred},\,L_{gt}) + \\ + \mathcal{L}_{1}(D_{pred},\,D_{gt}).
\end{multline}


To speed up and improve the quality of reconstruction, optimization of gaussians is performed on input images which resolution is consistently increasing ($H_g \times W_g$). The resolution of the input images is reduced by using a Gaussian pyramid. This approach allows to quickly set fairly accurate parameters for gaussians, which ensures fast convergence on large-dimensional images.

\subsection{Segmentation by Text Query Using DINOv2 Features}
\label{sec:segmentation_requests}

To perform segmentation based on a text query, our system employs the Talk2DINO model, which converts text embeddings from CLIP space into the embedding space of \mbox{DINOv2}. This transformation is necessary because CLIP and DINOv2 embeddings are trained on different tasks and exist in distinct vector spaces. Talk2DINO bridges this gap, allowing textual descriptions to be linked with visual features of the scene, enabling accurate open-vocabulary segmentation.

In our approach, an embedding map of size \(H \times W \times K\) is created by projecting the reconstructed 3D scene onto the current camera viewpoint. This projection is computed based on the contributions of all Gaussians in the scene to each pixel of the image. 

The segmentation process consists of several steps:
\begin{enumerate}
    \item \textbf{Text Embedding Generation:} The text query is converted into an embedding of size \(1 \times 768\) using the CLIP model.
    \item \textbf{Feature Space Alignment:} Talk2DINO transforms this embedding into the DINOv2 feature space, producing a vector of size \(1 \times 768\).
    \item \textbf{Dimensionality Reduction:} PCA is applied to compress the text embedding to \(1 \times K\), aligning it with the visual embeddings from the \(H \times W \times K\) map. This reduces computational overhead during similarity comparison.
    \item \textbf{Semantic Similarity Computation:} The embedding map \(H \times W \times K\), generated via scene projection, is used to compute a similarity map with the text query. This is achieved by performing a dot product between each scene embedding and the compressed text embedding. The result is a semantic similarity map of size \(H \times W \times 1\).
\end{enumerate}

By projecting the reconstructed scene through Gaussians onto a 2D plane, the system generates an up-to-date semantic feature map for each frame, which is then used to segment objects based on the query.
\section{Experiments}
\label{sec:experiments}

\begin{table*}[t!]
\centering
\small
\begin{tabular}{l|c|c|c|c|c|c|c}
\hline
\textbf{Method} & \textbf{ATE RMSE ↓} & \textbf{Depth L1 ↓} & \textbf{FPS ↑} & \textbf{PSNR (dB) ↑} & \textbf{SSIM ↑} & \textbf{LPIPS ↓} & \textbf{Semantics} \\ \hline
NICE-SLAM~\cite{zhu_nice-slam_2022}  & 2.51  & 1.903  & 0.198  & 24.42  & 0.809  & 0.233  & /  \\
Vox-Fusion~\cite{yang_vox-fusion_2022} & 1.47  & 2.913  & 1.07   & 24.41  & 0.801  & 0.236  & /  \\
Co-SLAM~\cite{wang_co-slam_2023}    & 1.06  & 1.513  & 6.41   & 30.24  & 0.939  & 0.252  & /  \\
ESLAM~\cite{johari_eslam_2023}       & 0.62  & 0.945  & 3.02   & 29.08  & 0.929  & 0.239  & /  \\
SplaTAM~\cite{keetha_splatam_nodate} & 0.41  & 0.49  & 0.21   & 34.11  & 0.968  & 0.102  & /  \\
Gaussian-SLAM~\cite{yugay_gaussian-slam_2024} & \textbf{0.31} & 0.68  & 0.63  & 42.08  & \textbf{0.996} & \textbf{0.018} &  / \\
MonoGS~\cite{matsuki_gaussian_2024}  & 0.79  & / & 0.445 & 37.50  & 0.96  & 0.07  & /  \\ 
NEDS-SLAM~\cite{ji_neds-slam_2024}  & 0.354  & 0.47 & - & 34.76 & 0.962 & 0.088  & Closed-Vocabulary  \\
RGBDS-SLAM~\cite{cao_rgbds-slam_2024}   & 0.589  & \textbf{0.342} & - & 38.85 & 0.967 & 0.035  & Closed-Vocabulary  \\
SNI-SLAM~\cite{zhu_sni-slam_2024}   & 0.456 & 0.766  & 2.15 & - & - & -  & Closed-Vocabulary  \\
GS$^3$SLAM~\cite{li_gs3_2024}  & 0.37 & -  & - & 36.26 & 0.989 & 0.052  & Closed-Vocabulary  \\ 
SGS-SLAM~\cite{li_sgs-slam_2025} & 0.412 & 0.356  & 2.11 & 34.15 & 0.97  & 0.096   & Closed-Vocabulary  \\ 
OVO-Gaussian-SLAM~\cite{martins_ovo-slam_2024} & \textbf{0.31} & 0.68  & 0.28 & \textbf{42.08} & \textbf{0.996} & \textbf{0.018} & Open-Vocabulary    \\ \hline
Photo-SLAM~\cite{huang_photo-slam_2024} (baseline) & 1.24  & /  & \textbf{27.35}  & 30.91  & 0.883  & \textbf{0.143}  & /  \\ 
LEG-SLAM (Ours)                        & \textbf{0.94}  & \textbf{2.15}  & 10.56  & \textbf{32.12}  & \textbf{0.9151} & 0.1456 & Open-Vocabulary    \\ \hline
\end{tabular}%

\caption{Quantitative comparison of SLAM accuracy, speed, rendering quality, and semantic capabilities for LEG-SLAM and other SLAM methods. The results are averaged over 8 scenes from the Replica dataset.}
\label{tab:replica_slam_comparison}
\end{table*}
In this section, we conduct a series of experiments to evaluate the effectiveness of LEG-SLAM in the task of open-vocabulary 3D scene reconstruction with semantic understanding. We analyze the method's performance on various datasets and compare it with existing approaches.

First, we evaluate the quality of 3D reconstruction on the Replica dataset. Then, we assess semantic segmentation accuracy on ScanNet dataset, demonstrating open-vocabulary scene understanding. Next, we analyze SLAM performance on Replica, examining the accuracy of camera trajectory estimation and tracking speed. Additionally, we investigate the impact of different factors, including embedding compression methods, visual feature extraction architectures, and rendering strategies.

Finally, we compare LEG-SLAM with existing methods in terms of 3D reconstruction quality, semantic segmentation accuracy, and computational efficiency, highlighting its advantages for real-time language-enhanced SLAM.

\begin{figure*}[th!]
    \centering
    \begin{subfigure}{0.24\textwidth}
        \centering
        \includegraphics[width=\linewidth]{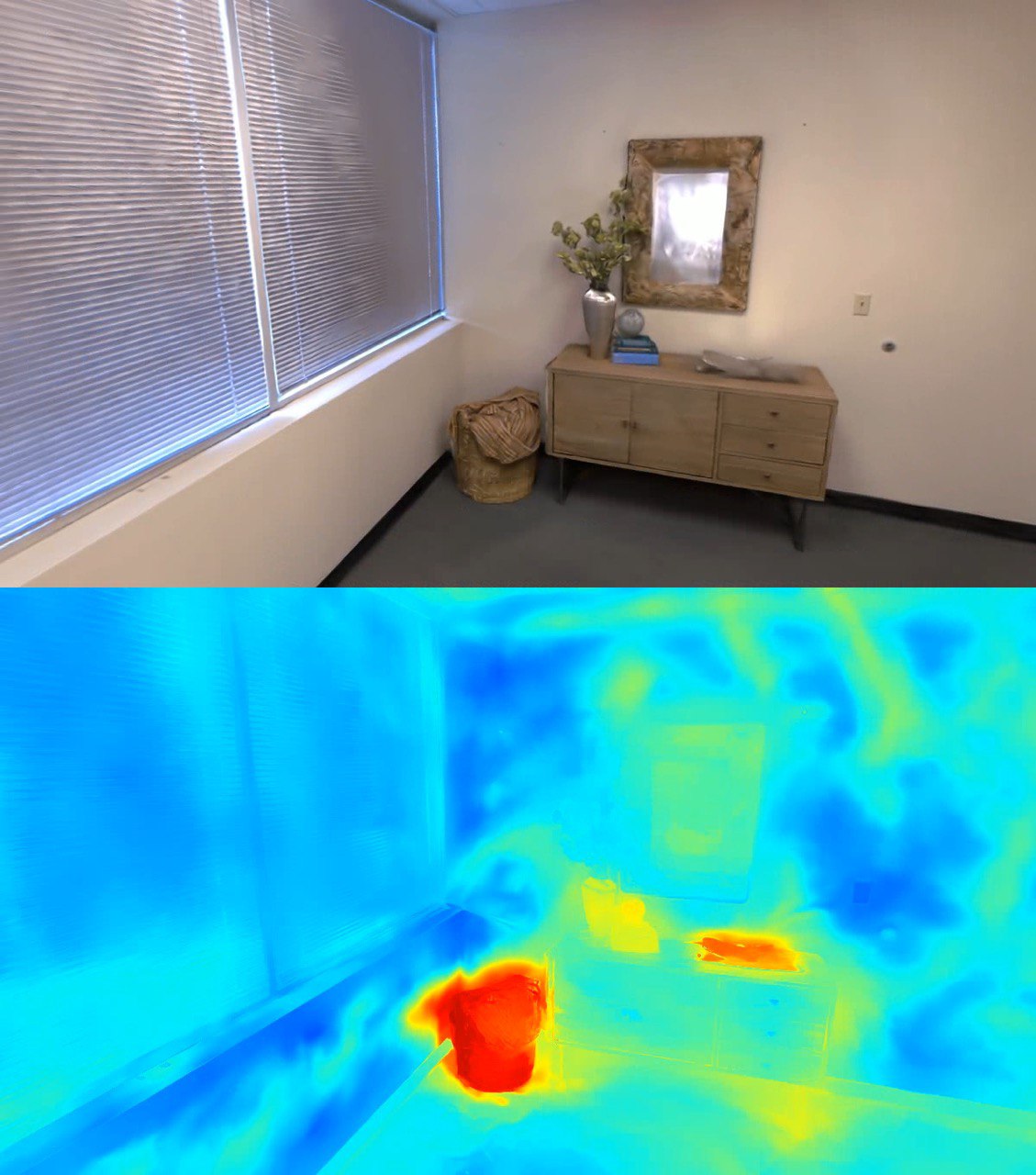}
        \caption{Text query: "basket"}
        \label{fig:image_a}
    \end{subfigure}
    \begin{subfigure}{0.24\textwidth}
        \centering
        \includegraphics[width=\linewidth]{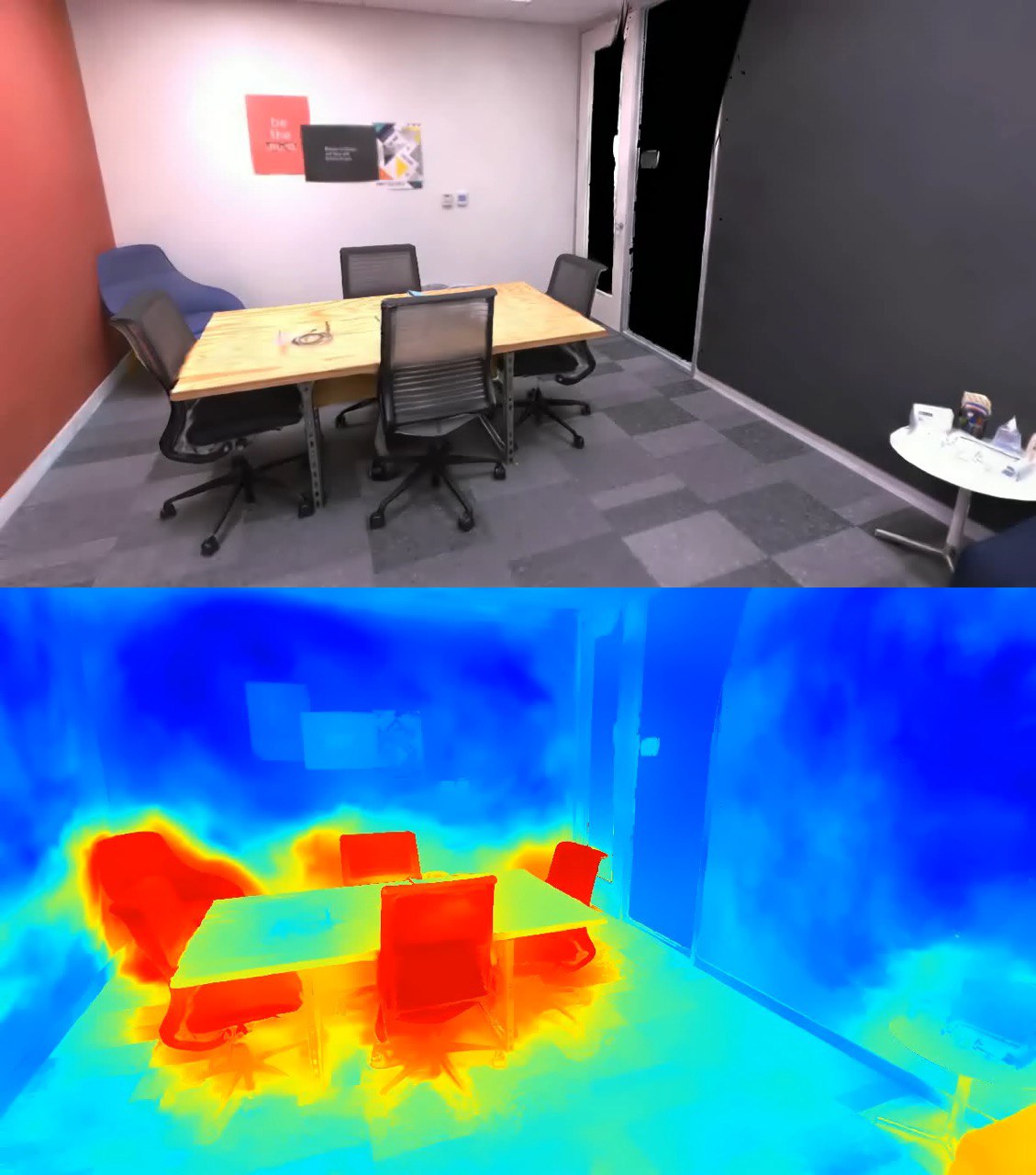}
        \caption{Text query: "chair"}
        \label{fig:image_b}
    \end{subfigure}
    \begin{subfigure}{0.24\textwidth}
        \centering
        \includegraphics[width=\linewidth]{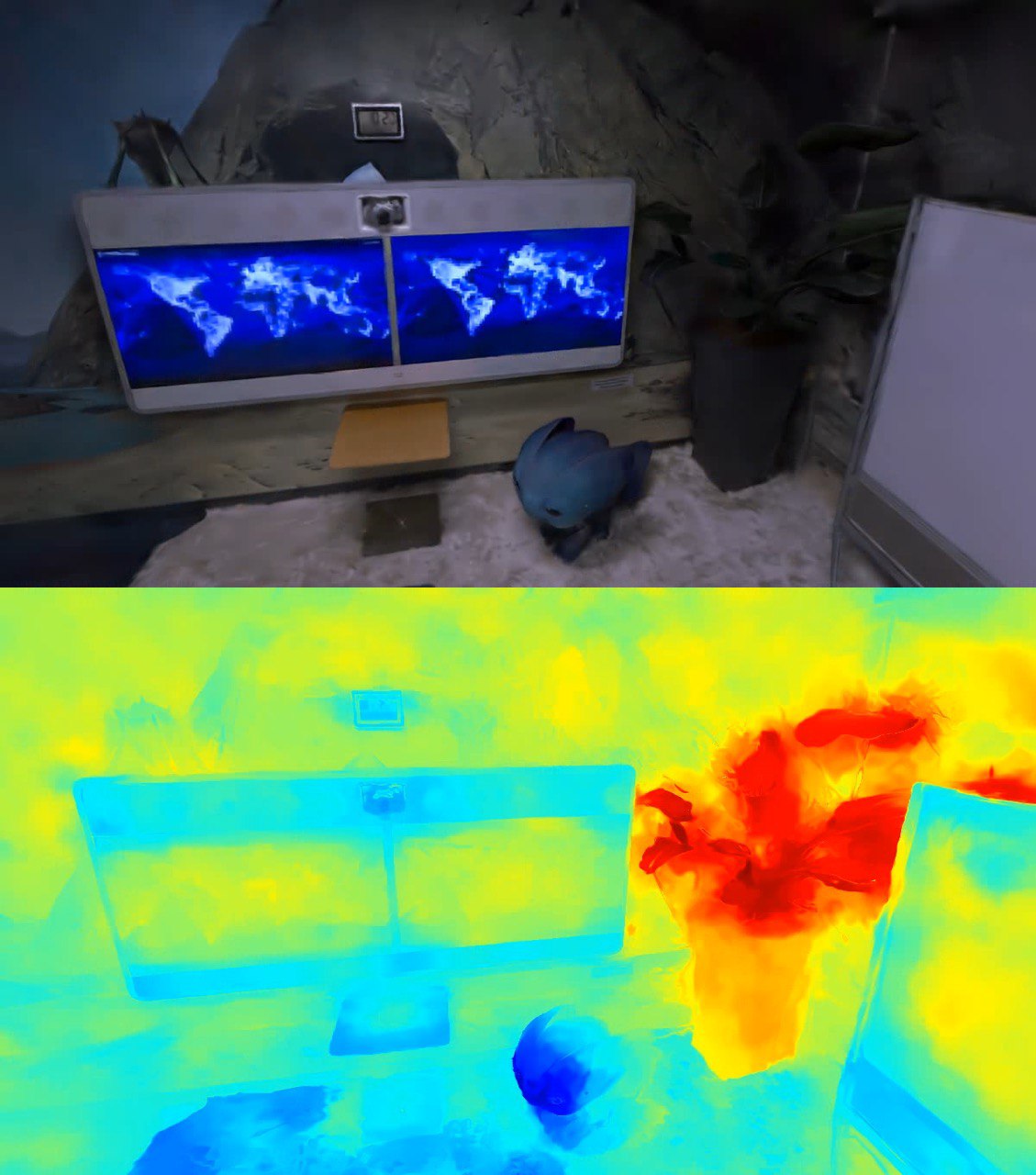}
        \caption{Text query: "plant"}
        \label{fig:image_c}
    \end{subfigure}
    \begin{subfigure}{0.24\textwidth}
        \centering
        \includegraphics[width=\linewidth]{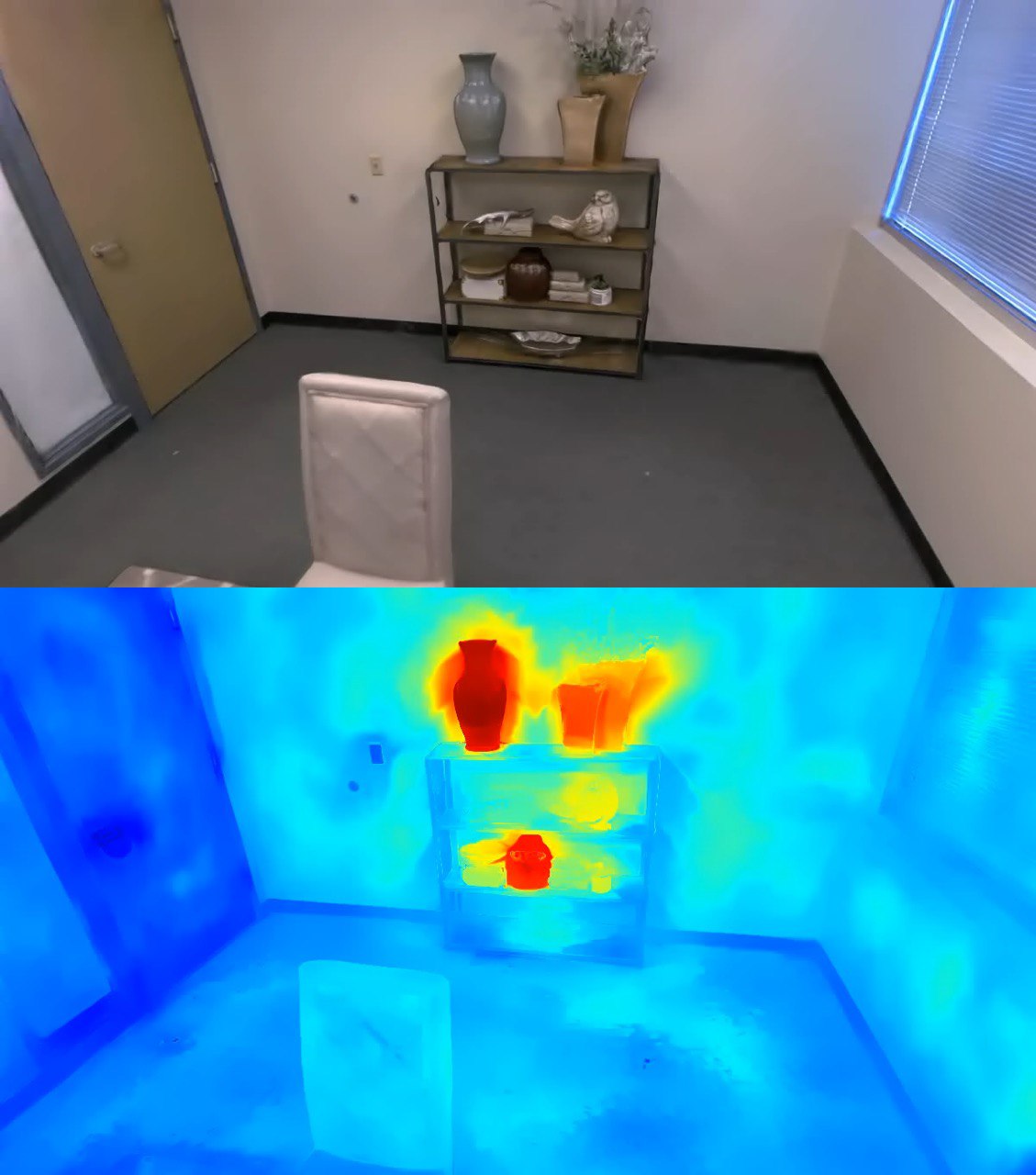}
        \caption{Text query: "vase"}
        \label{fig:image_d}
    \end{subfigure}
    
    \caption{Visualization of RGB and semantic heatmap outputs for different text queries. Each image contains both an RGB rendering (top) and a semantic heatmap (bottom) highlighting regions matching the given query.}
    \label{fig:quality_figure}
\end{figure*}

\subsection{Experimental Setup}
\label{sec:experimental_setup}

\subsubsection{Datasets}
To evaluate the quality of scene reconstruction and semantic segmentation in an open-vocabulary setting, we use several widely adopted datasets in 3D Gaussian Splatting research.

Replica \cite{straub_replica_2019} is a high-quality synthetic dataset containing realistic 3D-rendered indoor environments. It is widely used for evaluating 3D reconstruction methods, as it provides scenes with detailed geometry and textures. In our experiments, Replica serves as a benchmark for assessing the accuracy of scene reconstruction.

ScanNet \cite{dai_scannet_2017} is an RGB-D dataset consisting of real-world scanned indoor scenes with annotated 3D camera poses and 2D object class labels. Due to the presence of semantic annotations, it is well-suited for evaluating semantic segmentation performance. Our experiments are conducted on 12 validation scenes from ScanNet.

Each of these datasets plays a distinct role in our evaluation. While Replica provides a controlled environment for assessing reconstruction quality, ScanNet allows us to analyze both reconstruction and semantic segmentation in real-world indoor scenes. This diverse selection ensures a comprehensive assessment of LEG-SLAM across multiple challenging scenarios, covering both 3D reconstruction and semantic scene understanding.

\subsubsection{Implementation Details}
\label{sec:implementation_details}

All experiments were conducted on an NVIDIA RTX 4090 GPU with 24 GB of video memory. We use DINOv2 ViT-B/14 as the backbone for extracting visual features and CLIP ViT-B/16 for processing text queries. To project textual embeddings into the visual feature space, we employ the Talk2DINO module.

\subsection{Quantitative Analysis}

To further illustrate the effectiveness of our method in open-vocabulary semantic understanding, Figure~\ref{fig:quality_figure} presents RGB renderings and corresponding semantic heatmaps for various text queries across different scenes. Each image showcases how LEG-SLAM successfully localizes and highlights queried objects, even for ambiguous or spatially complex environments.

For instance, Figure~\ref{fig:image_a} demonstrates accurate identification of a “basket”, while figure~\ref{fig:image_b} highlights the correct localization of a “chair”. Figure~\ref{fig:image_c} showcases the method’s ability to recognize a “plant”, where fine-grained textures and occlusions make segmentation challenging. Finally, Figure~\ref{fig:image_d} presents a successful retrieval of a “vase”, capturing both its geometric structure and material properties.

These qualitative results confirm that LEG-SLAM not only provides high-fidelity 3D reconstructions but also accurately aligns semantic information with the rendered environment, making it well-suited for open-vocabulary object localization in real-time applications.

\subsection{Scene Reconstruction}
\label{sec:scene_reconstruction}

To evaluate the quality of 3D reconstruction, we conducted experiments on the Replica dataset. Table \ref{tab:replica_slam_comparison} presents the results of our method compared to existing SLAM approaches.

Our method achieves competitive reconstruction quality, slightly trailing behind SplaTAM in terms of PSNR and SSIM. This difference arises from the additional semantic rendering in our approach, which introduces constraints on the optimization process. Despite this, our method maintains strong perceptual quality, as reflected in LPIPS scores, demonstrating its effectiveness in jointly reconstructing geometry and semantics in real-time applications.

\subsection{Open-Vocabulary Semantic Segmentation}
\label{sec:semantic_segmentation}

We compare LEG-SLAM against state-of-the-art methods on the ScanNet dataset. The segmentation quality is evaluated using mIoU and mAcc metrics.

\begin{figure*}[htbp]
    \centering
    \begin{subfigure}{0.15\textwidth}
        \centering
        \textbf{LSeg}
    \end{subfigure}
    \begin{subfigure}{0.15\textwidth}
        \centering
        \textbf{Feature3DGS}
    \end{subfigure}
    \begin{subfigure}{0.15\textwidth}
        \centering
        \textbf{LangSplat}
    \end{subfigure}
    \begin{subfigure}{0.15\textwidth}
        \centering
        \textbf{Semantic Gaussians}
    \end{subfigure}
    \begin{subfigure}{0.15\textwidth}
        \centering
        \textbf{LEG-SLAM (Ours)}
    \end{subfigure}
    \begin{subfigure}{0.15\textwidth}
        \centering
        \textbf{GT}
    \end{subfigure}

    \begin{subfigure}{0.15\textwidth}
        \includegraphics[width=\linewidth]{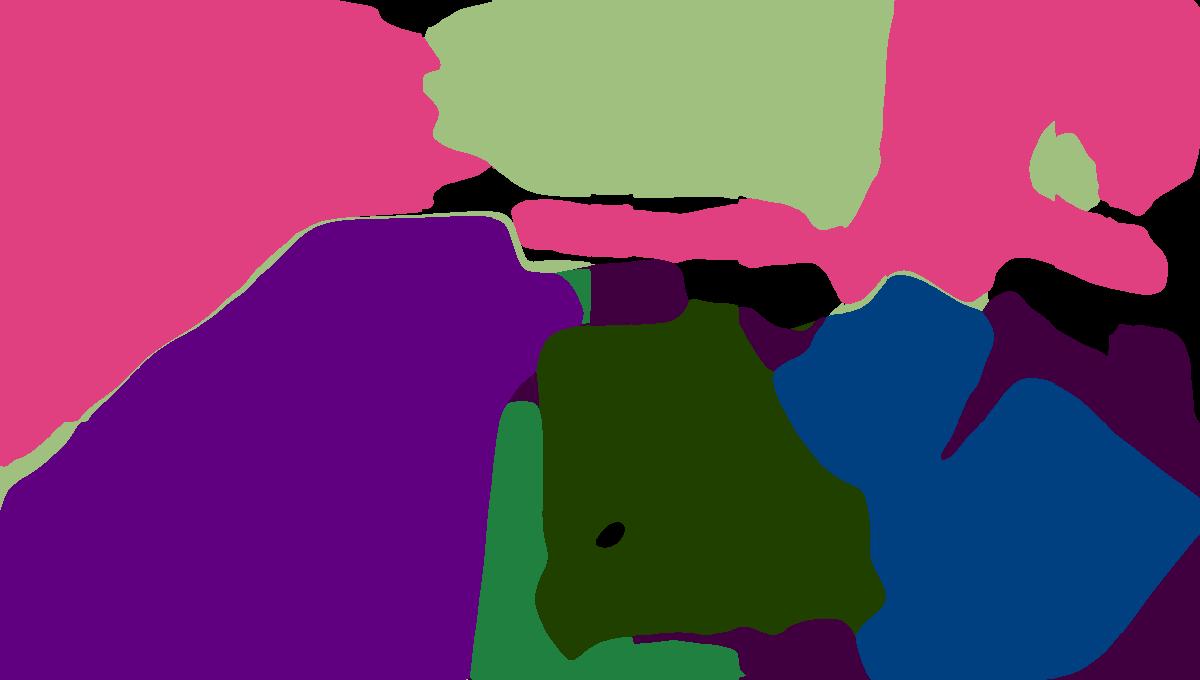}
    \end{subfigure}
    \begin{subfigure}{0.15\textwidth}
        \includegraphics[width=\linewidth]{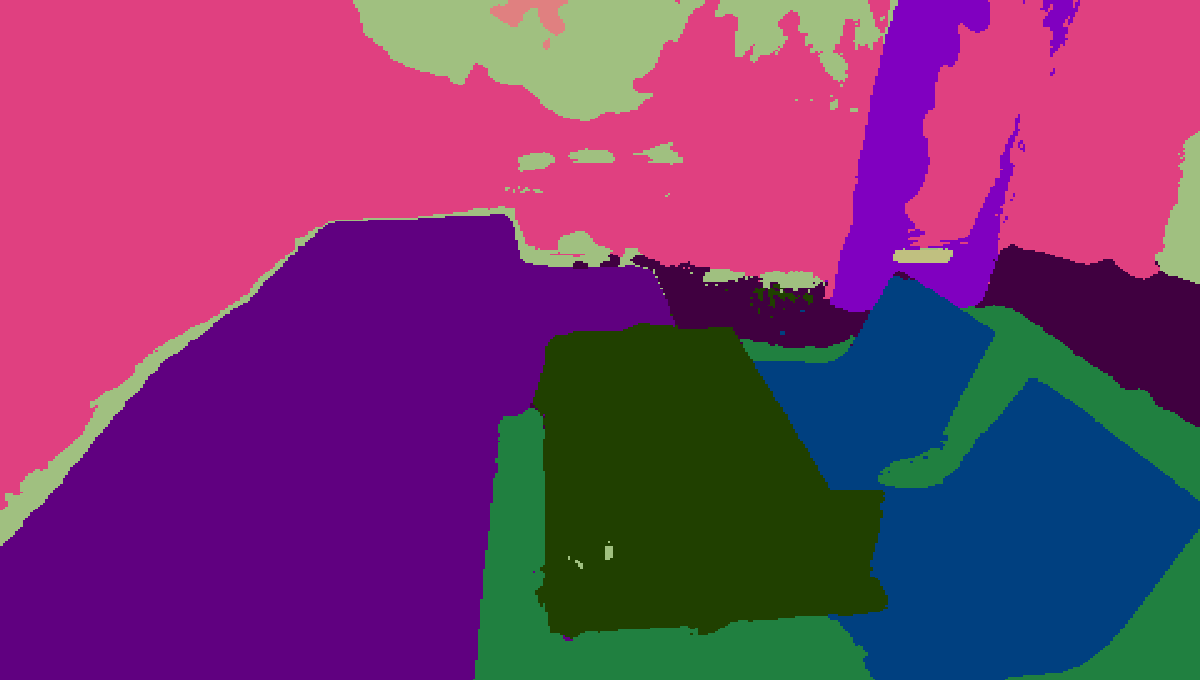}
    \end{subfigure}
    \begin{subfigure}{0.15\textwidth}
        \includegraphics[width=\linewidth]{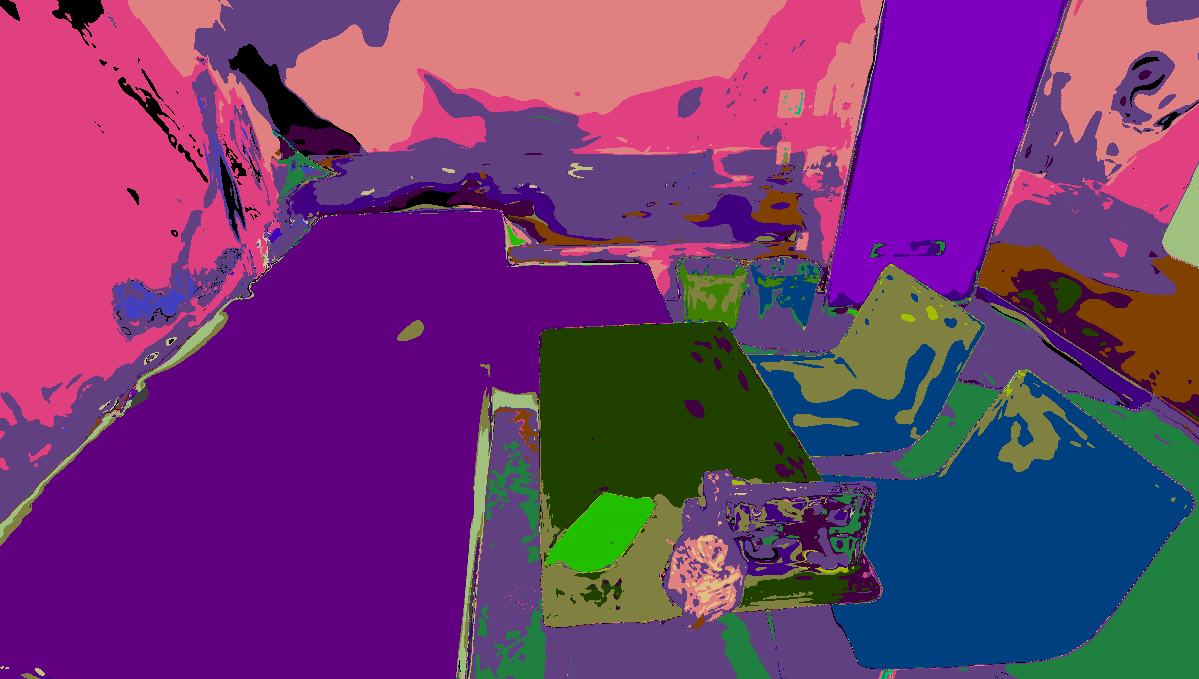}
    \end{subfigure}
    \begin{subfigure}{0.15\textwidth}
        \includegraphics[width=\linewidth]{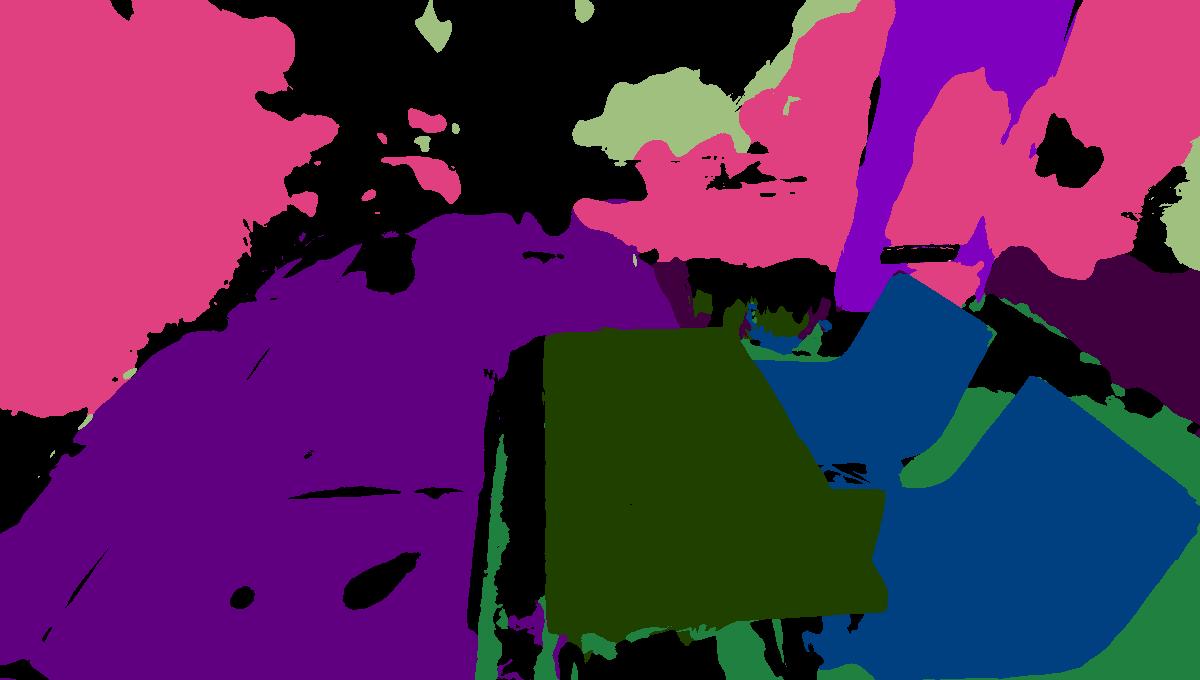}
    \end{subfigure}
    \begin{subfigure}{0.15\textwidth}
        \includegraphics[width=\linewidth]{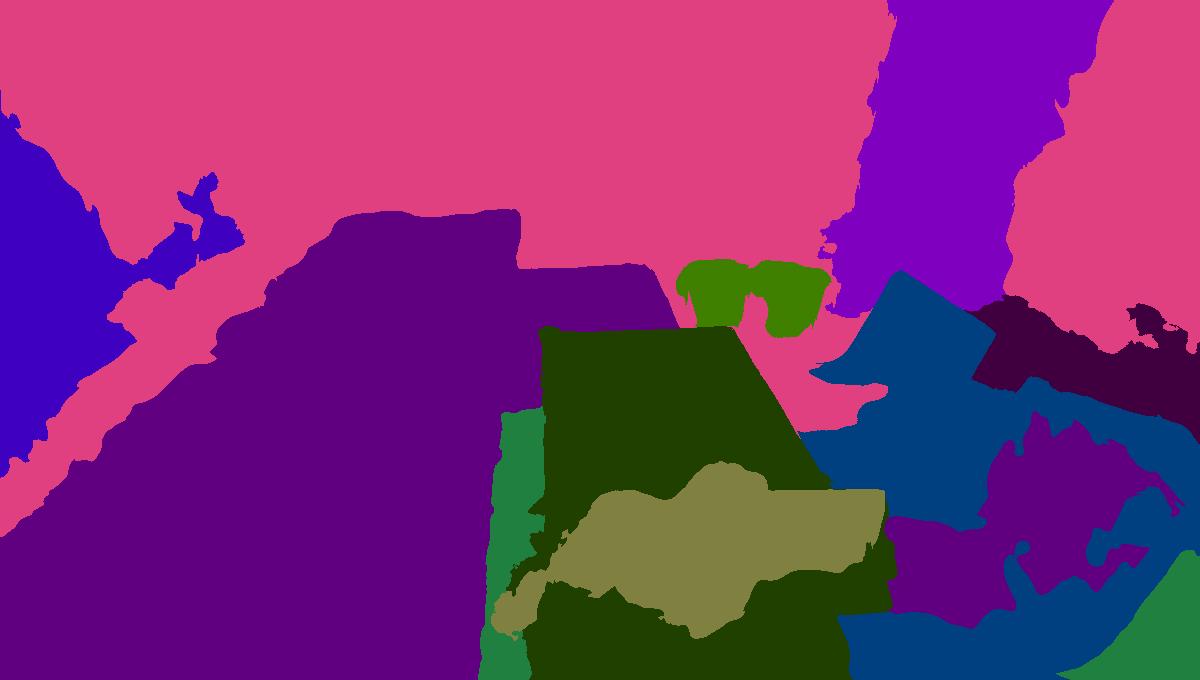}
    \end{subfigure}
    \begin{subfigure}{0.15\textwidth}
        \includegraphics[width=\linewidth]{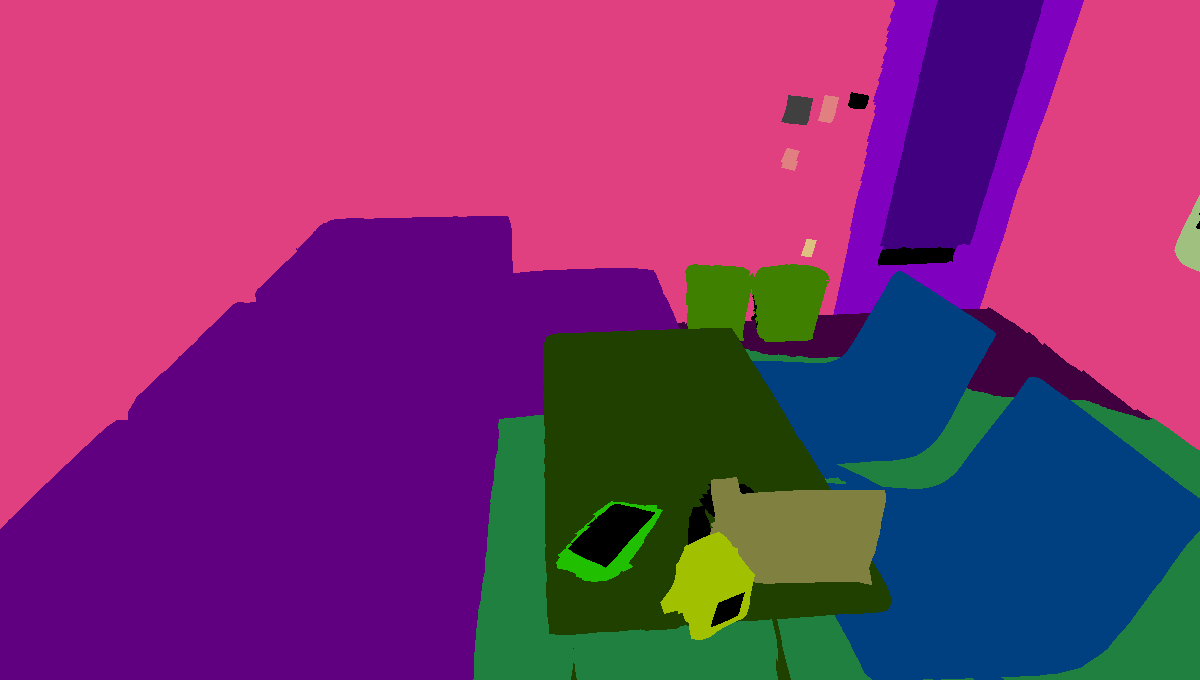}
    \end{subfigure}

    \begin{subfigure}{0.15\textwidth}
        \includegraphics[width=\linewidth]{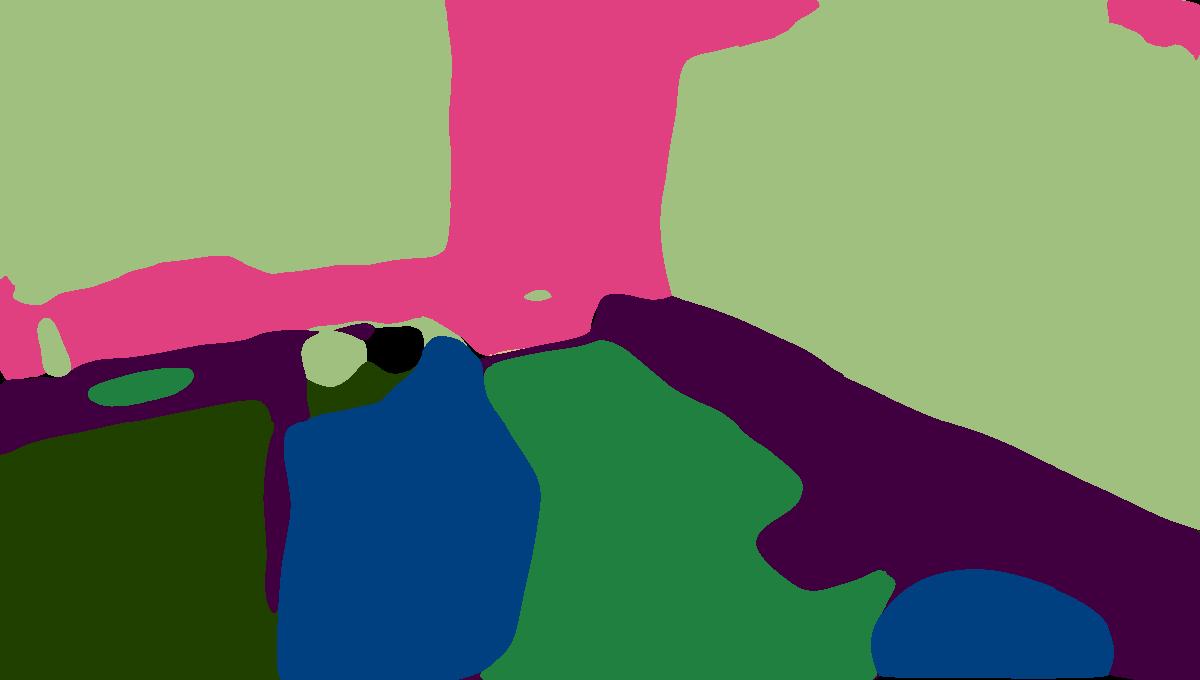}
    \end{subfigure}
    \begin{subfigure}{0.15\textwidth}
        \includegraphics[width=\linewidth]{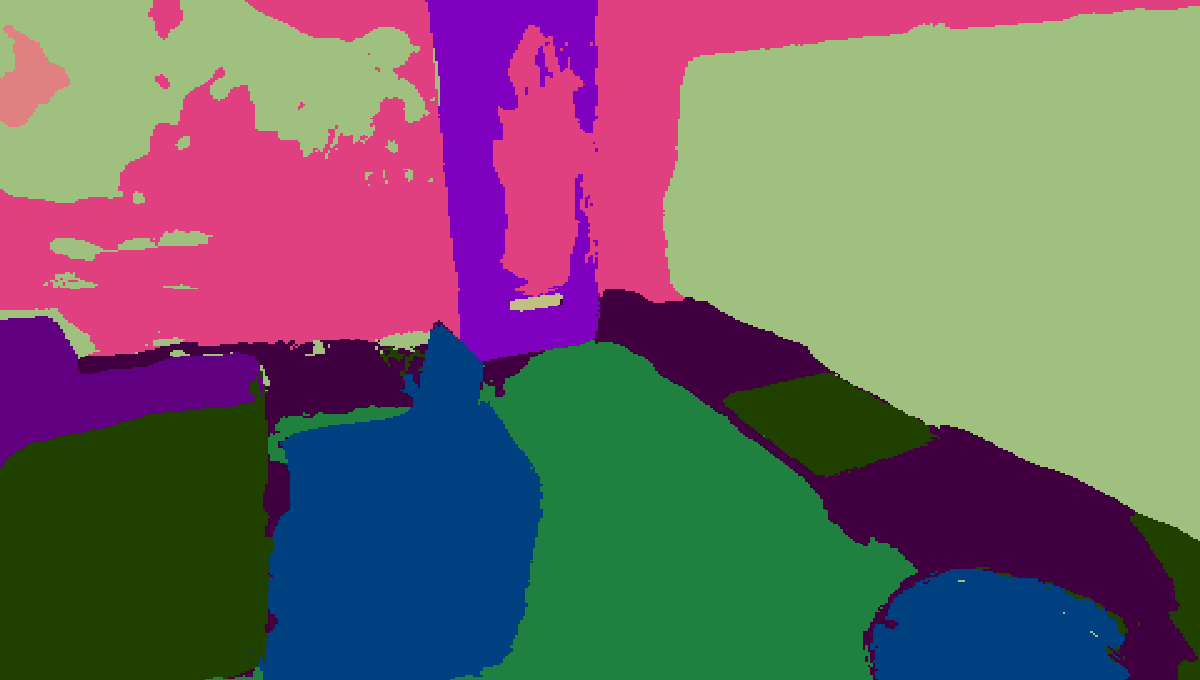}
    \end{subfigure}
    \begin{subfigure}{0.15\textwidth}
        \includegraphics[width=\linewidth]{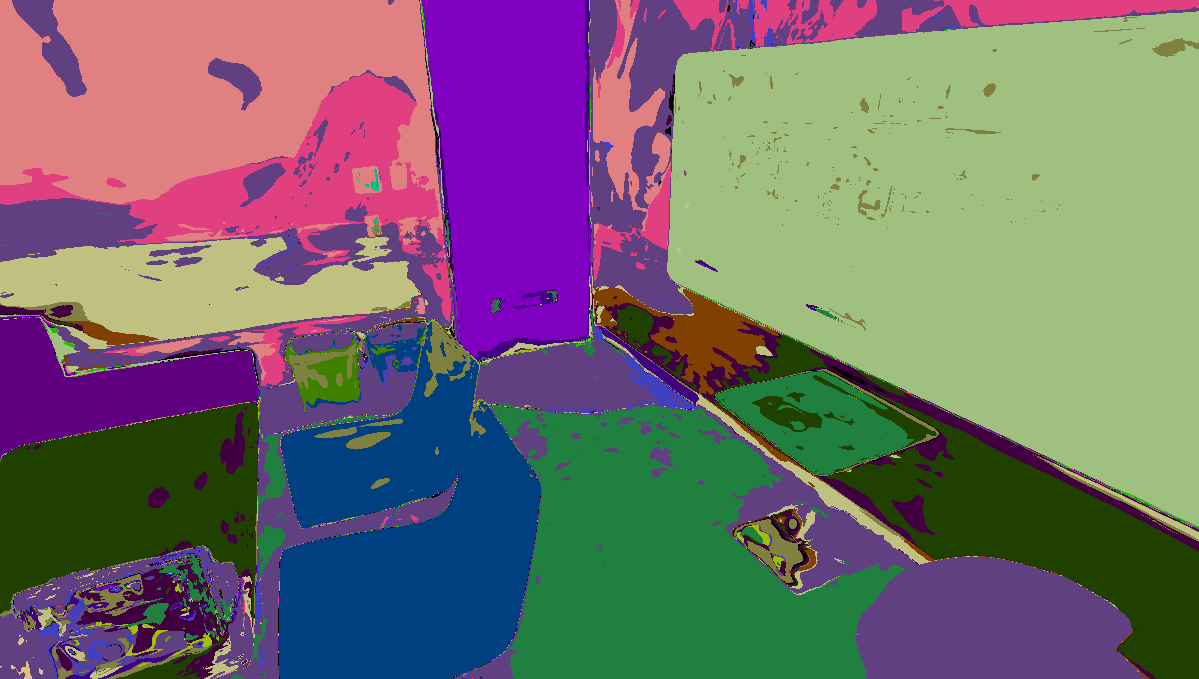}
    \end{subfigure}
    \begin{subfigure}{0.15\textwidth}
        \includegraphics[width=\linewidth]{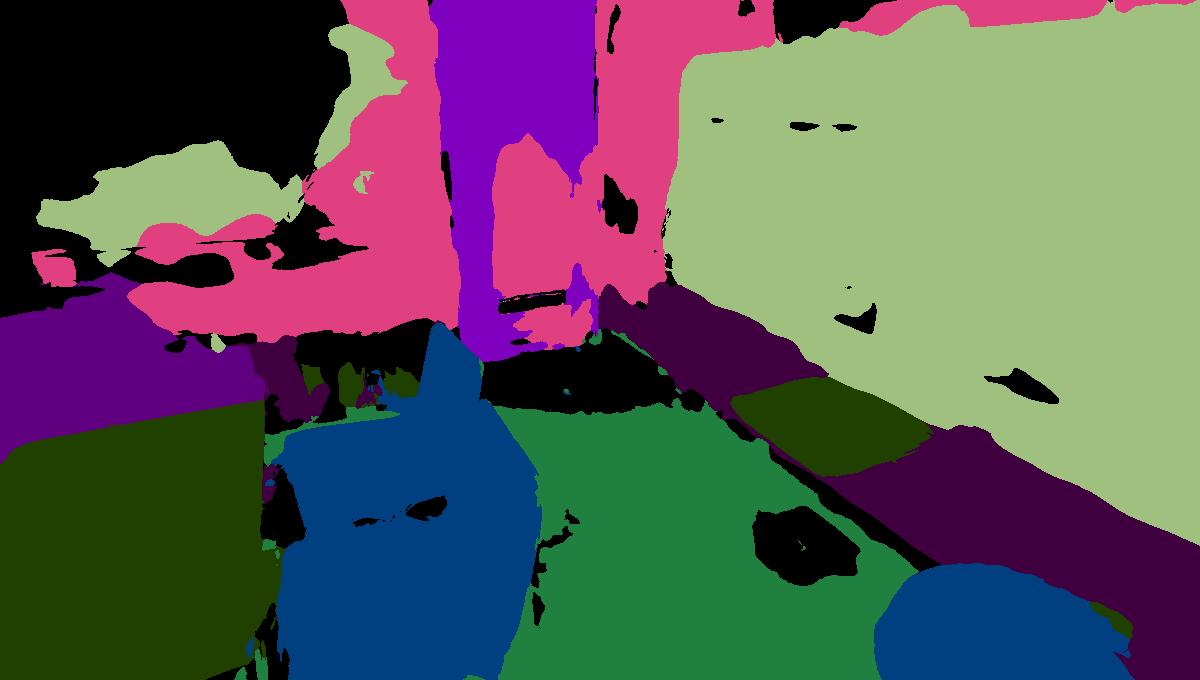}
    \end{subfigure}
    \begin{subfigure}{0.15\textwidth}
        \includegraphics[width=\linewidth]{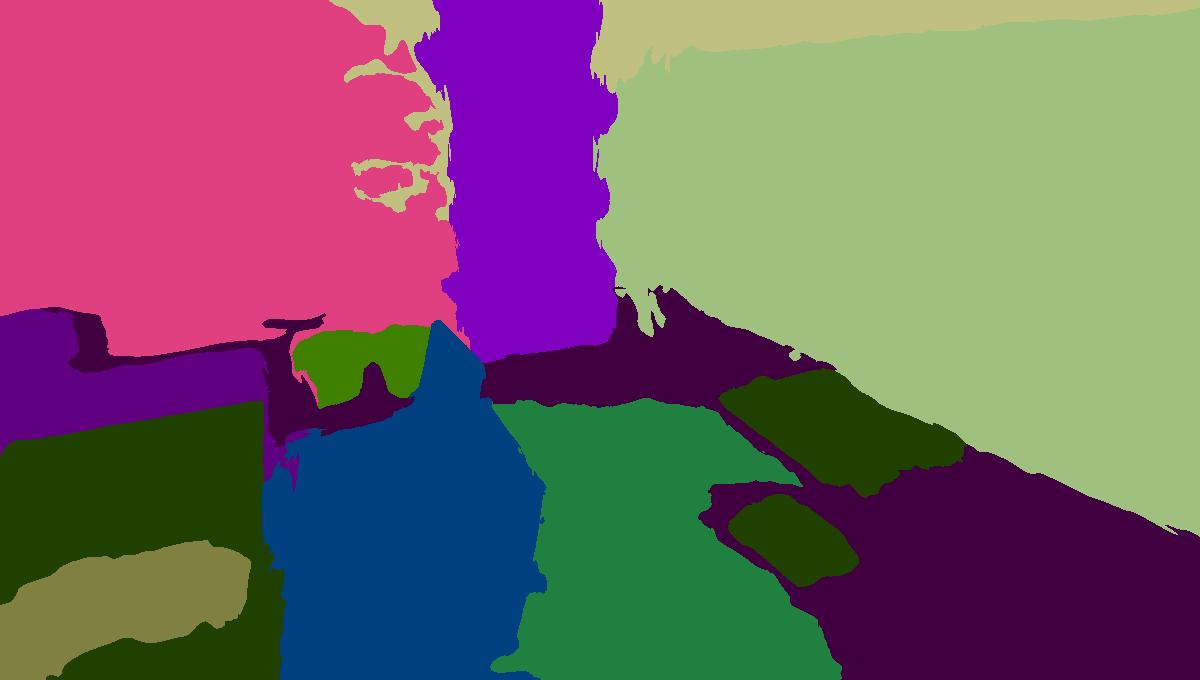}
    \end{subfigure}
    \begin{subfigure}{0.15\textwidth}
        \includegraphics[width=\linewidth]{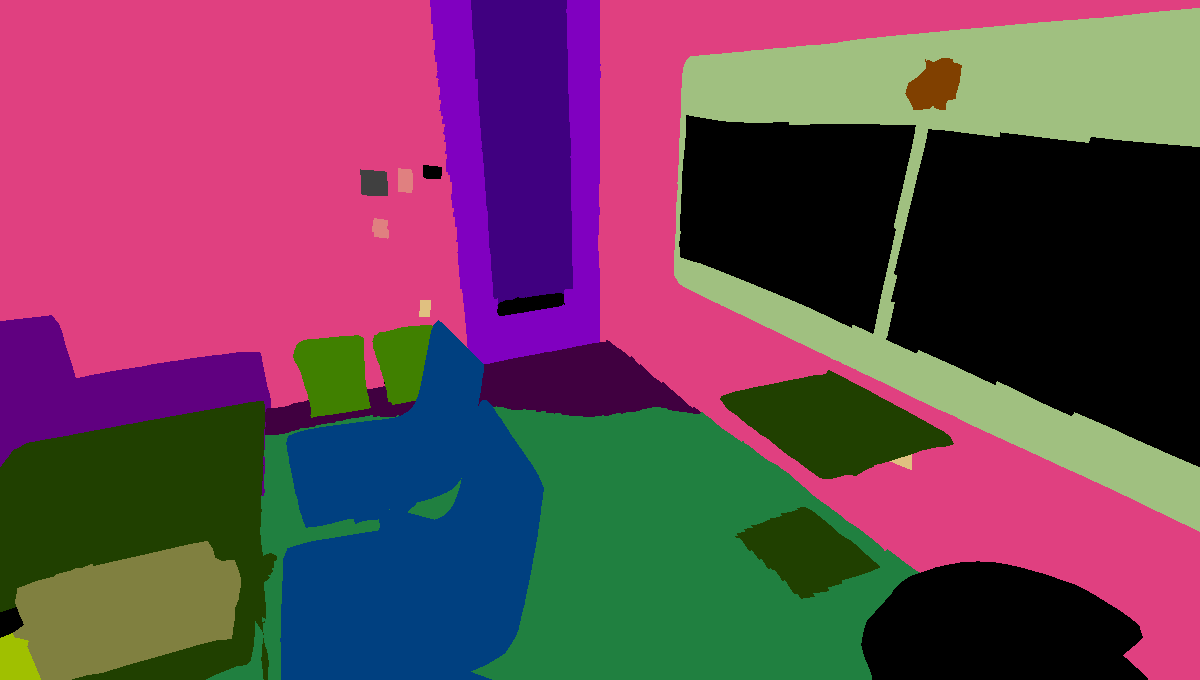}
    \end{subfigure}

    \begin{subfigure}{0.15\textwidth}
        \includegraphics[width=\linewidth]{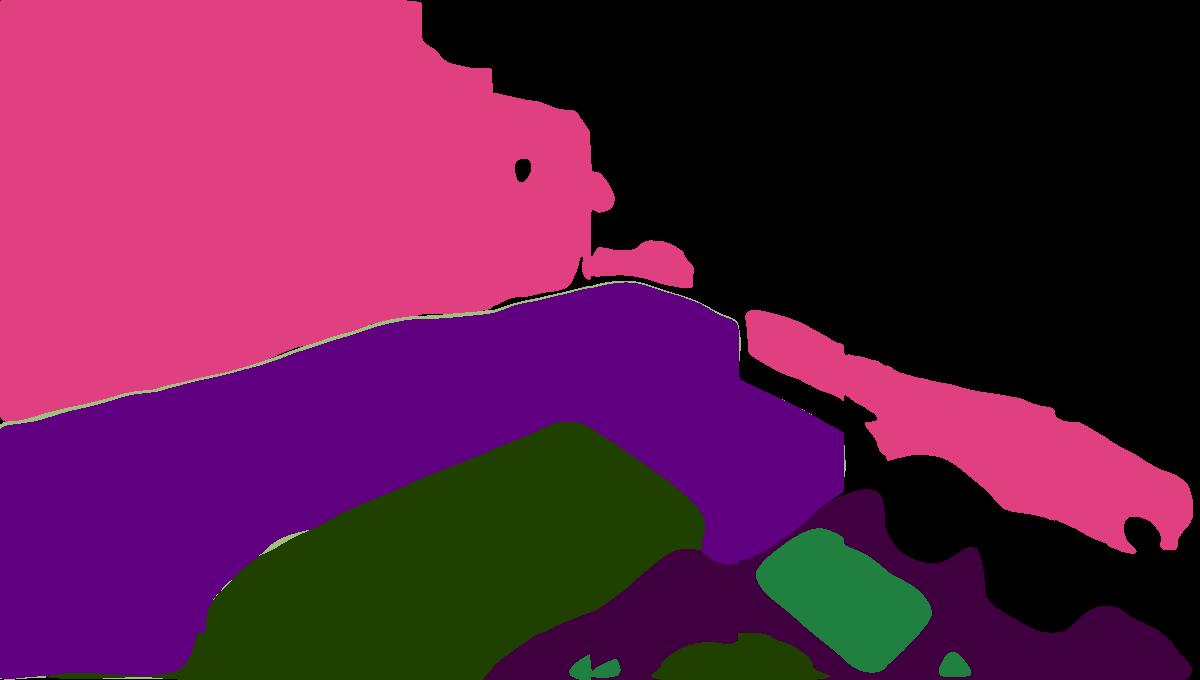}
    \end{subfigure}
    \begin{subfigure}{0.15\textwidth}
        \includegraphics[width=\linewidth]{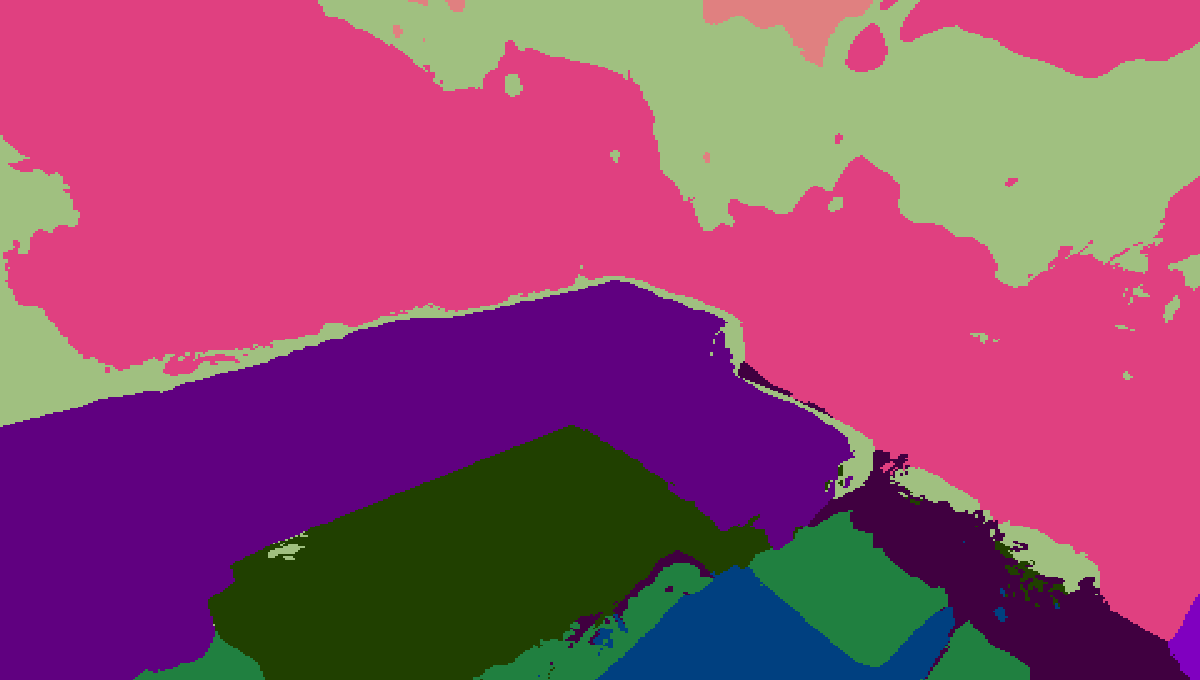}
    \end{subfigure}
    \begin{subfigure}{0.15\textwidth}
        \includegraphics[width=\linewidth]{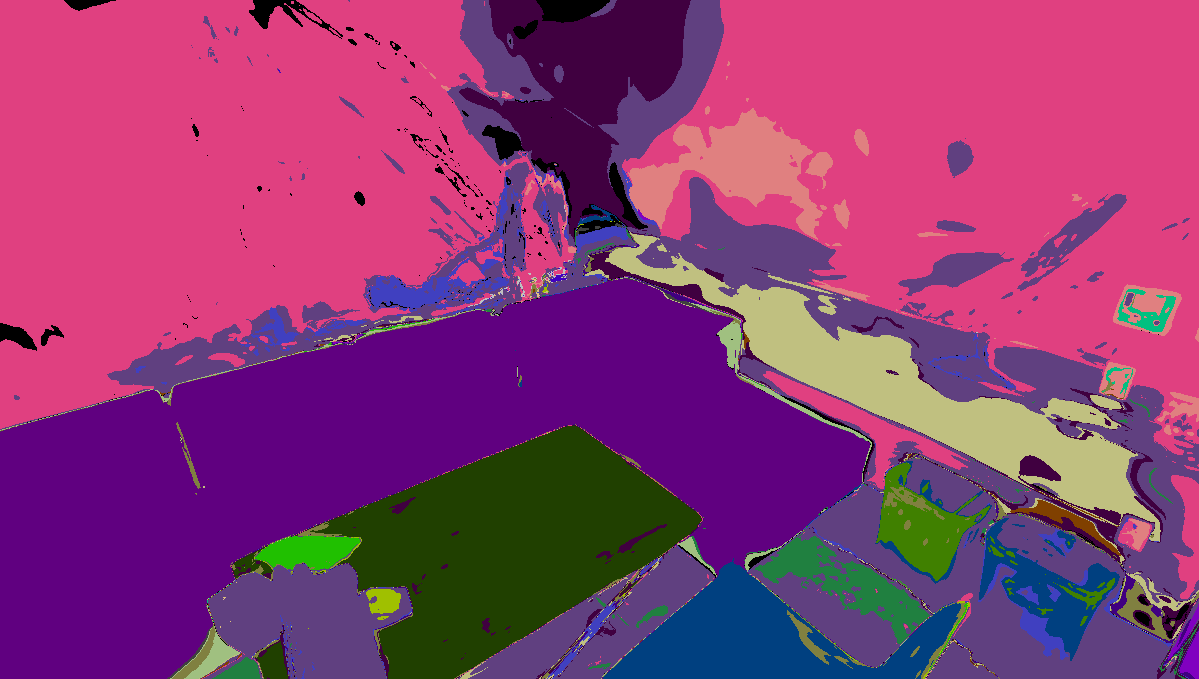}
    \end{subfigure}
    \begin{subfigure}{0.15\textwidth}
        \includegraphics[width=\linewidth]{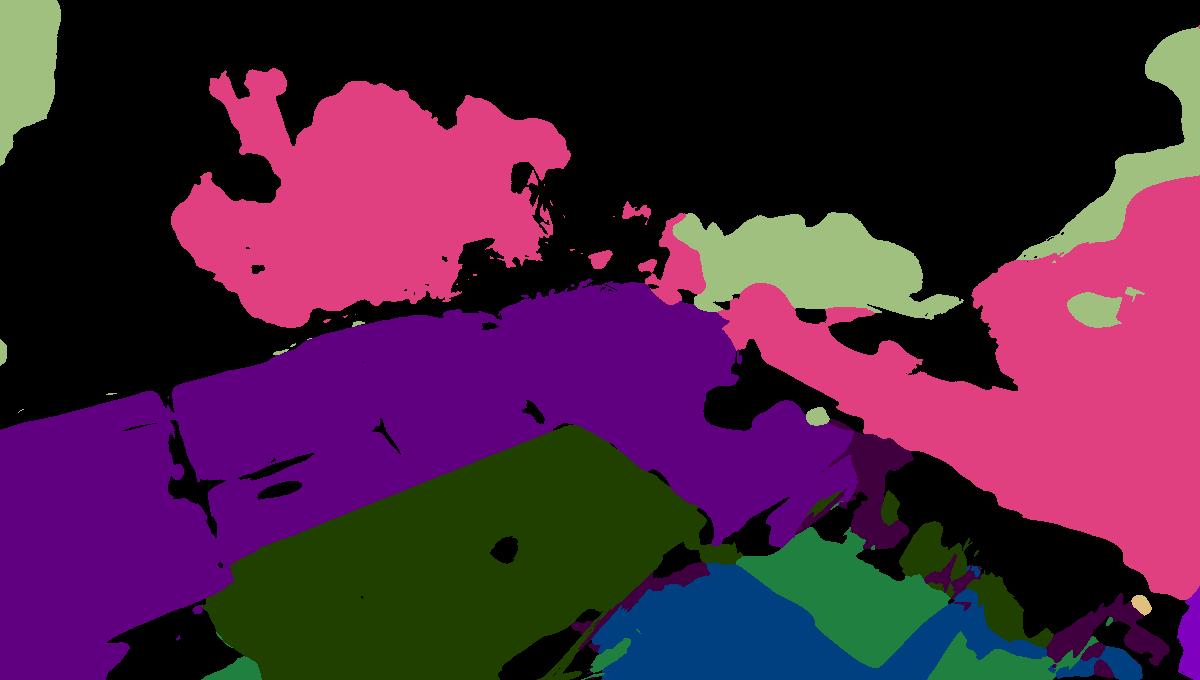}
    \end{subfigure}
    \begin{subfigure}{0.15\textwidth}
        \includegraphics[width=\linewidth]{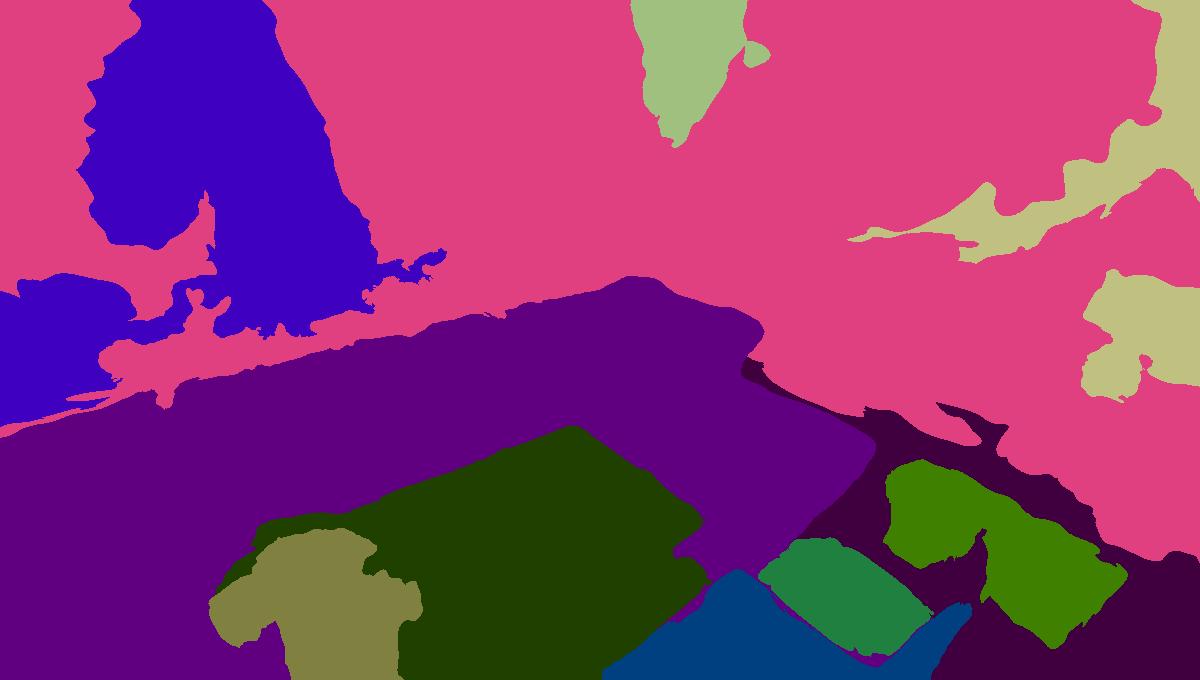}
    \end{subfigure}
    \begin{subfigure}{0.15\textwidth}
        \includegraphics[width=\linewidth]{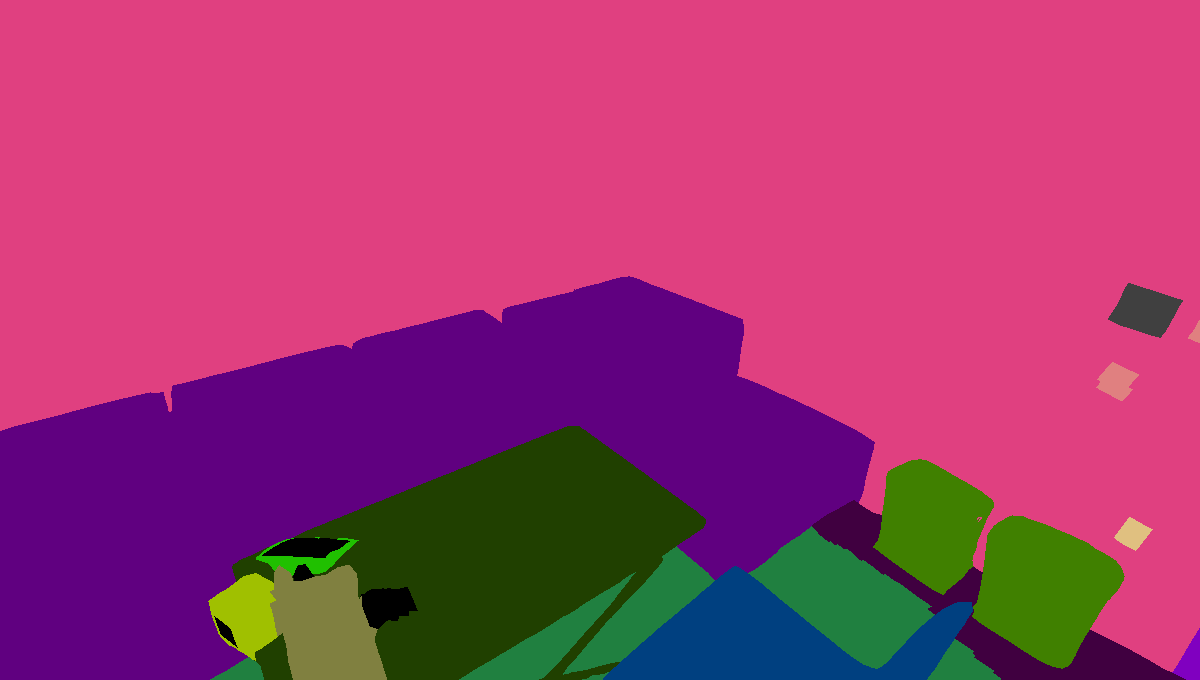}
    \end{subfigure}

    \begin{subfigure}{0.15\textwidth}
        \includegraphics[width=\linewidth]{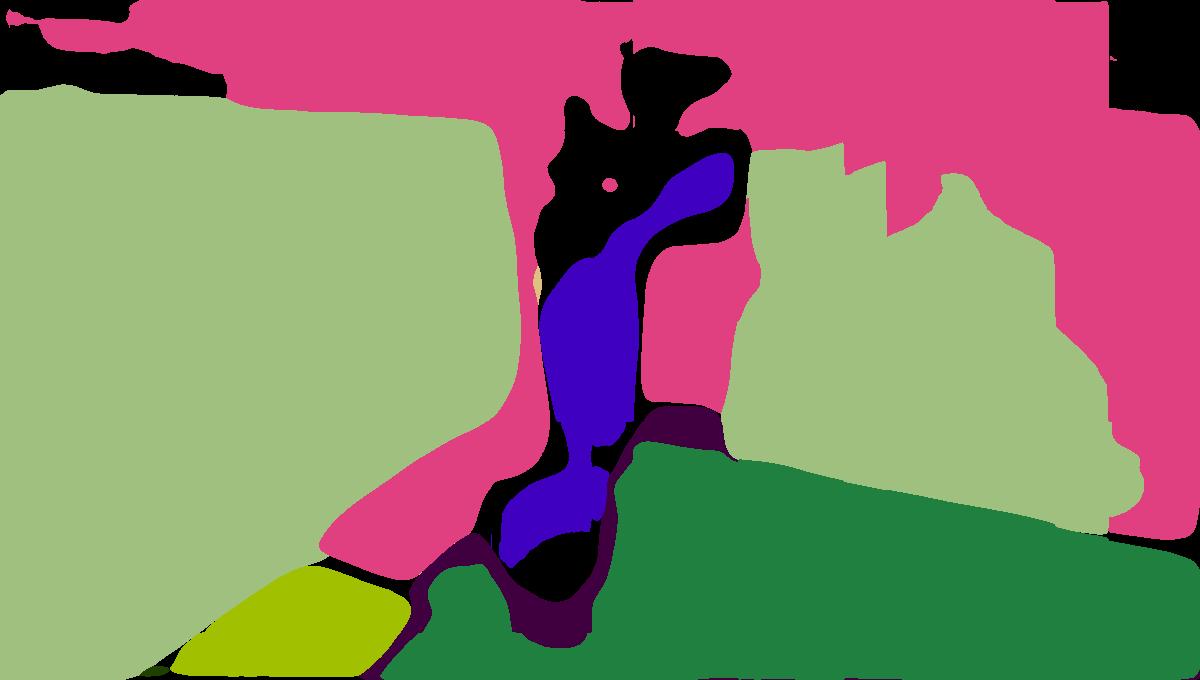}
    \end{subfigure}
    \begin{subfigure}{0.15\textwidth}
        \includegraphics[width=\linewidth]{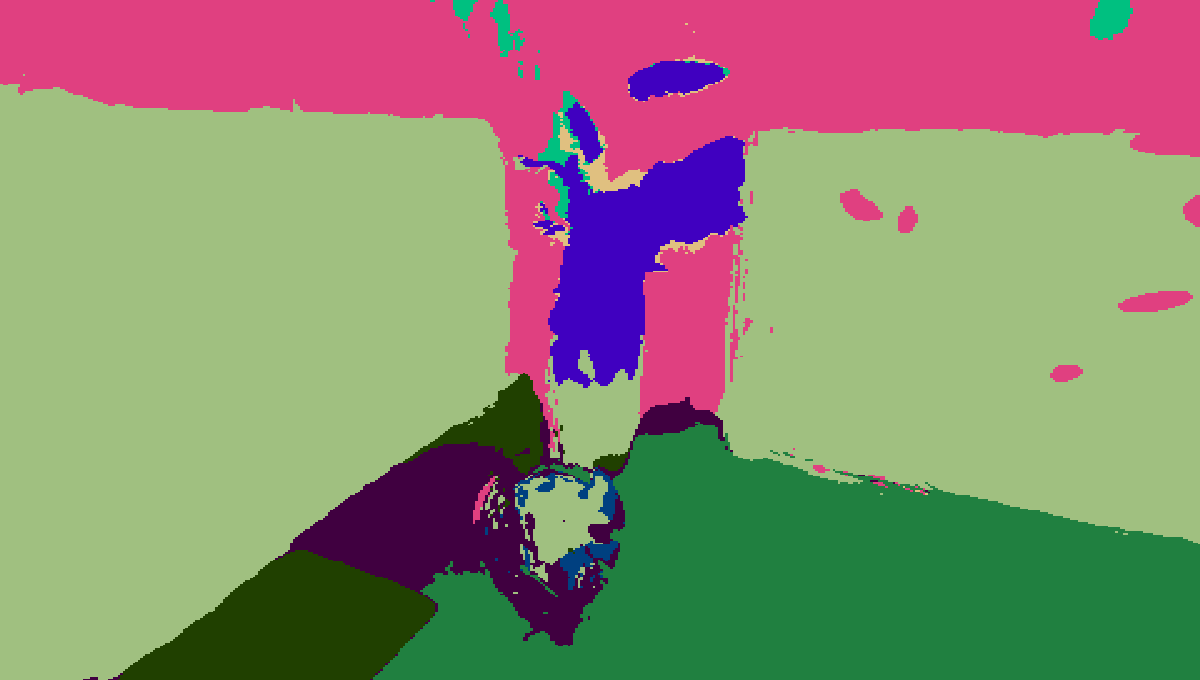}
    \end{subfigure}
    \begin{subfigure}{0.15\textwidth}
        \includegraphics[width=\linewidth]{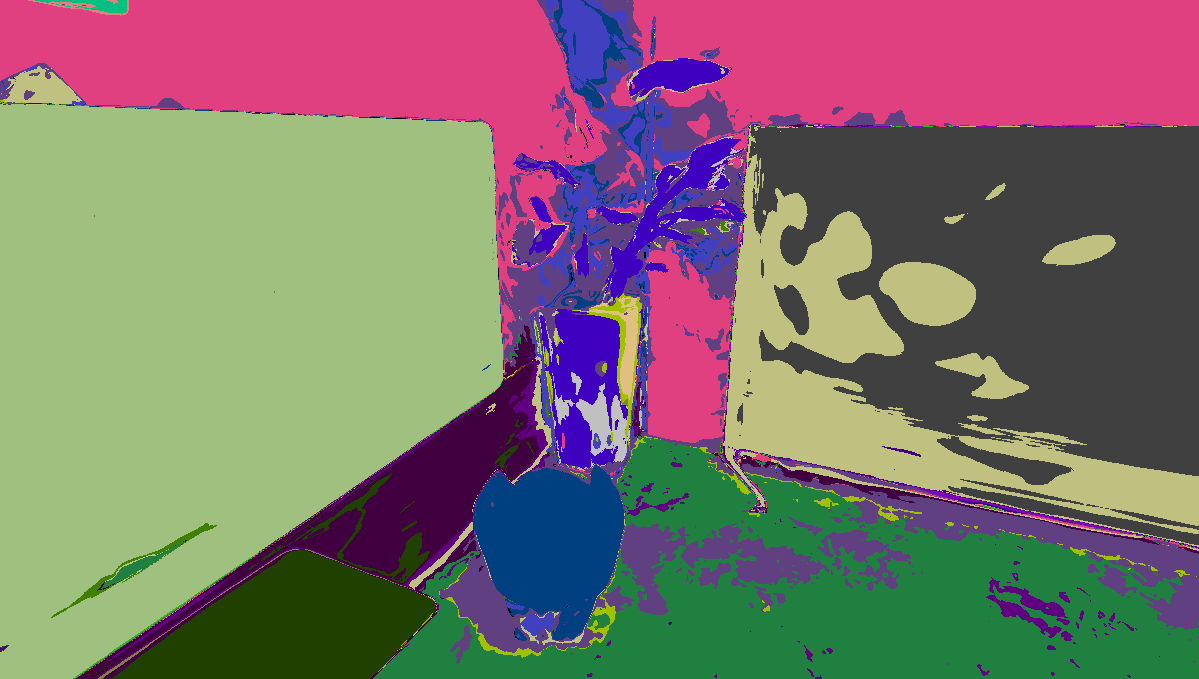}
    \end{subfigure}
    \begin{subfigure}{0.15\textwidth}
        \includegraphics[width=\linewidth]{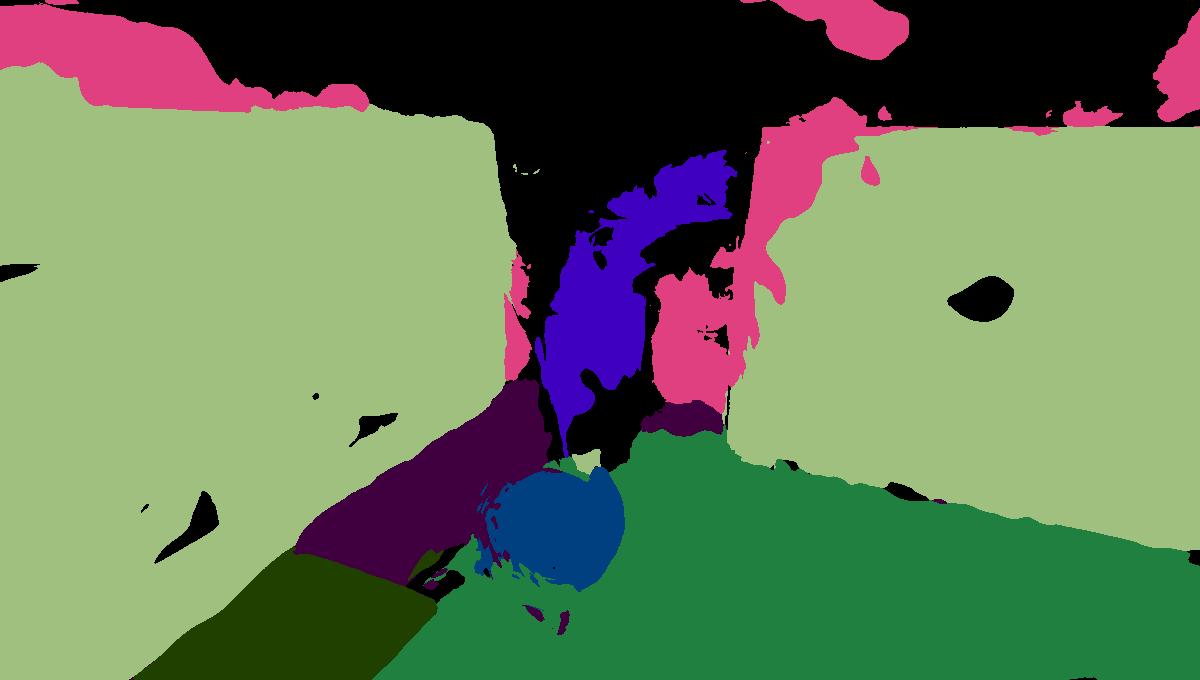}
    \end{subfigure}
    \begin{subfigure}{0.15\textwidth}
        \includegraphics[width=\linewidth]{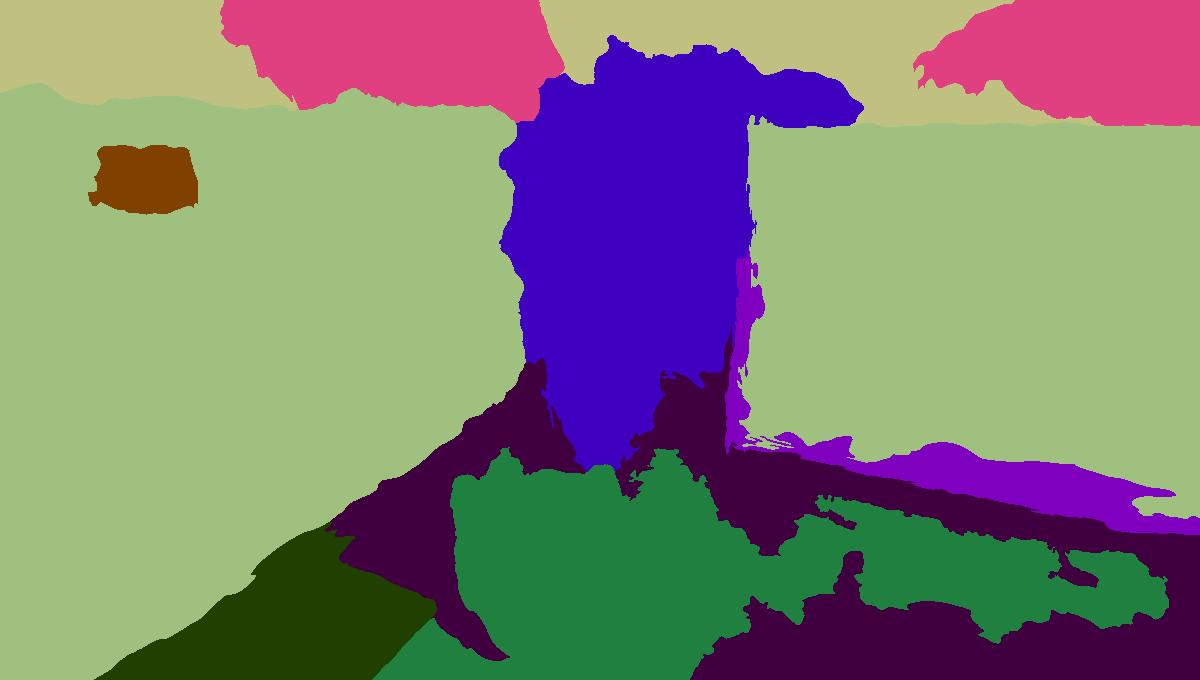}
    \end{subfigure}
    \begin{subfigure}{0.15\textwidth}
        \includegraphics[width=\linewidth]{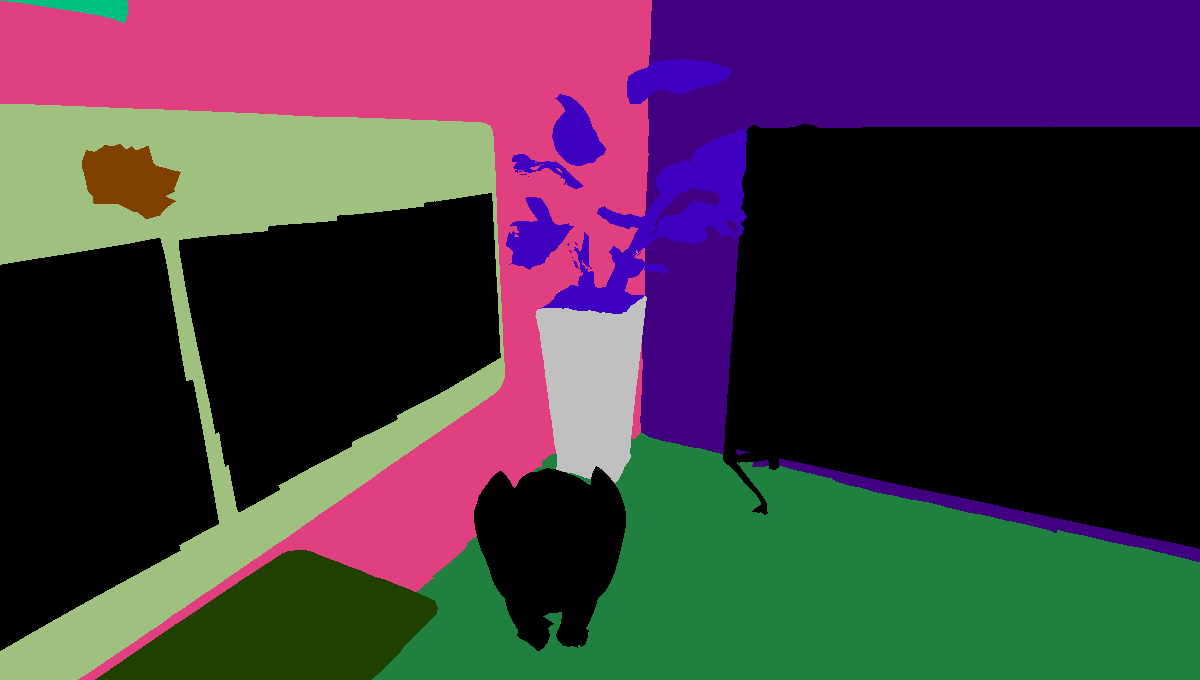}
    \end{subfigure}

    \caption{Comparison of semantic segmentation performance across different pipelines on the Replica dataset, including ground truth for reference.}
    \label{fig:replica_comparison_grid}
\end{figure*}

Table \ref{tab:3d_segmentation_results} presents a comparative analysis of NeRF and Gaussian Splatting-based methods, reporting the average segmentation metrics and training time across 12 ScanNet scenes. Figure \ref{fig:replica_comparison_grid} visually illustrates the segmentation results for different methods.

\begin{table}[ht]
\centering
\scriptsize
\begin{tabular}{p{2.3cm}|p{1.55cm}|p{0.6cm}|p{0.6cm}|p{1.0cm}}
\hline
\textbf{Method}          & \textbf{Backbone} & \textbf{mIoU, \%} & \textbf{mAcc, \%} & \textbf{Training Time} \\ \hline
LERF               & NeRF+CLIP   & 31.2          & 61.7          & 45 min \\
PVLFF              & NeRF+LSeg   & 52.9          & 67.0          & 65 min  \\
LangSplat      & 3DGS+CLIP   & 24.7          & 42.0          & 180 min \\
Feature3DGS  & 3DGS+LSeg   & 59.2          & 75.1          & 150 min  \\
OVO-Gaussian-SLAM  & 3DGS+CLIP  & 29.3 & 41.1 & 138 min  \\
OVO-mapping  & - & 38.1 & 50.5 & 2.6 min \\
Semantic Gaussians & 3DGS+LSeg   & \textbf{62.0}          & \textbf{77.0}          & 90 min \\
\textbf{LEG-SLAM}       & 3DGS+DINOv2 & 41.4 & 74.3 & \textbf{1.5 min} \\ \hline
\end{tabular}%

\caption{Comparison of Open-Vocabulary segmentation methods on 12 Scenes of ScanNet (Average 1800 Frames per Scene)}
\label{tab:3d_segmentation_results}
\end{table}

Results show that while LEG-SLAM does not achieve the highest mIoU, it significantly outperforms competing methods in runtime efficiency, running at 18 FPS on ScanNet. On average, it processes a scene in 1.5 minutes, whereas other methods require tens of minutes or even hours per scene.

\subsection{SLAM Comparison}
\label{sec:slam_comparison}

We evaluate the performance of LEG-SLAM in the SLAM task by comparing it with existing state-of-the-art methods on the Replica dataset. The primary metrics include ATE RMSE for trajectory accuracy and rendering quality. Additionally, we emphasize the system's speed, which is critical for real-time applications.

As shown in Table~\ref{tab:replica_slam_comparison}, LEG-SLAM achieves competitive reconstruction accuracy while significantly outperforming other methods in terms of speed, processing frames at 10 FPS. While SplaTAM yields slightly better reconstruction and rendering quality, it does not support real-time performance. The added complexity of optimizing 3D Gaussians for both geometric and semantic consistency introduces minor reconstruction trade-offs but enables open-vocabulary semantic segmentation, making LEG-SLAM the only method in this comparison capable of integrating real-time SLAM with open-vocabulary scene understanding.

\subsection{Ablation studies}

\subsubsection{Comparison of Different Feature Extraction Backbones}

Selecting an efficient feature extraction method is crucial for real-time performance. We compare two architectures: DINOv2 ViT-B/14 and OpenSeg. Both provide high-quality visual features but differ in computational efficiency and adaptability to open-vocabulary tasks.

DINOv2 is a self-supervised model trained to produce generalizable visual embeddings without relying on predefined categories. This makes it well-suited for diverse environments, where it maintains high feature informativeness. In contrast, OpenSeg is trained on COCO categories and performs well in segmentation but struggles with generalization in open-vocabulary settings.

To evaluate computational efficiency, we measured processing time, including embedding compression using PCA. Results show that OpenSeg + PCA takes 100 ms, while DINOv2 + PCA completes the same operation in 33 ms — a 3× speedup. Given the constraints of real-time operation, this significant reduction in latency makes DINOv2 the preferred choice, offering a balance between feature quality and computational cost.

\begin{table}[ht]
\centering
\small
\begin{tabular}{p{2.5cm}|c|c|c}
\hline
\textbf{Method} & \textbf{CosSim $\uparrow$} & \textbf{mIoU $\uparrow$} & \textbf{Time (ms) $\downarrow$} \\ \hline
MLP-AE (2-layer) & 0.9379 & 37.3 & 0.85 \\ 
MLP-AE (4-layer) & 0.9551 & 34.8 & 2.96 \\ 
\textbf{PCA} & \textbf{0.9570} & \textbf{41.2} & \textbf{0.60} \\ \hline
\end{tabular}%
\caption{Comparison of embedding compression methods (all methods compress to 64 dimensions)}
\label{tab:compression_comparison}
\end{table}

\subsubsection{Influence of Embedding Compression Methods}

Efficient embedding compression is crucial for real-time applications. Table~\ref{tab:compression_comparison} compares PCA and two MLP-based Autoencoders (MLP-AE), all compressing embeddings from 768 to 64 dimensions.

PCA achieves the best trade-off between accuracy and efficiency, yielding the highest cosine similarity and segmentation quality while maintaining the fastest inference time. The 2-layer MLP-AE shows a noticeable drop in segmentation accuracy, while the deeper 4-layer variant improves reconstruction quality but suffers from significantly higher computational cost. Despite its deeper architecture, the 4-layer model does not outperform PCA, making it less suitable for real-time deployment.

Given these results, PCA emerges as the most effective choice, offering a superior balance between accuracy, computational speed, and implementation simplicity.

\begin{table}[t]
\centering
\label{tab:pca_metrics}
\small
\begin{tabular}{c|c|c|c|c}
\hline
\textbf{Dim} & \textbf{Cos. Sim.$\uparrow$}  & \textbf{mIoU$\uparrow$} & \textbf{PSNR$\uparrow$} & \textbf{Train FPS$\uparrow$} \\ \hline
3   & 0.7423  & 6.6  & 24.34  & 18.38 \\ 
6   & 0.8348  & 20.1  & \textbf{25.07}  & \textbf{18.81} \\ 
9   & 0.8814  & 27.2  & 24.83  & 18.72 \\ 
12  & 0.9070  & 32.0  & 24.53  & 18.63 \\ 
16  & 0.9298  & 37.1  & 24.7  & 18.4 \\ 
24  & 0.9532  & 39.1  & 24.75  & 18.22 \\ 
32  & 0.9662  & 40.1  & 24.48  & 18.09 \\ 
48  & 0.9808  & 40.4  & 24.30  & 17.52 \\ 
64  & 0.9880  & 41.4  & 24.26  & 17.55 \\ 
80  & 0.9920  & 42.2  & 23.96  & 17.63 \\ 
96  & 0.9946  & 42.3  & 23.65  & 17.58 \\ 
128  & \textbf{0.9973}  & \textbf{42.7}  & 23.57  & 17.44 \\ \hline
\end{tabular}
\caption{Results of PCA-based embedding compression experiments}
\end{table}

\subsubsection{Impact of embedding dimensionality}

To determine the optimal dimensionality for PCA-compressed embeddings, we conducted two separate evaluations: compression quality analysis using Cosine Similarity and full pipeline performance measured by mIoU, PSNR, and training speed in FPS.

For compression quality assessment, we randomly selected 10 frames from each of the 12 ScanNet validation scenes and extracted DINOv2 embeddings at the original resolution of 648 × 484 × 768. To prevent redundancy from neighboring pixels, embeddings were sampled with a stride of 4, ensuring a diverse set of semantic features. PCA was then applied with embedding dimensionalities ranging from 3 to 128, and reconstruction quality was assessed using Cosine Similarity.

To evaluate the overall impact on pipeline performance, the full LEG-SLAM pipeline was executed on complete ScanNet scenes, where segmentation quality was measured using mIoU, PSNR, and training FPS.

\begin{figure}[htbp]
    \centering
    \includegraphics[width=0.47\textwidth]{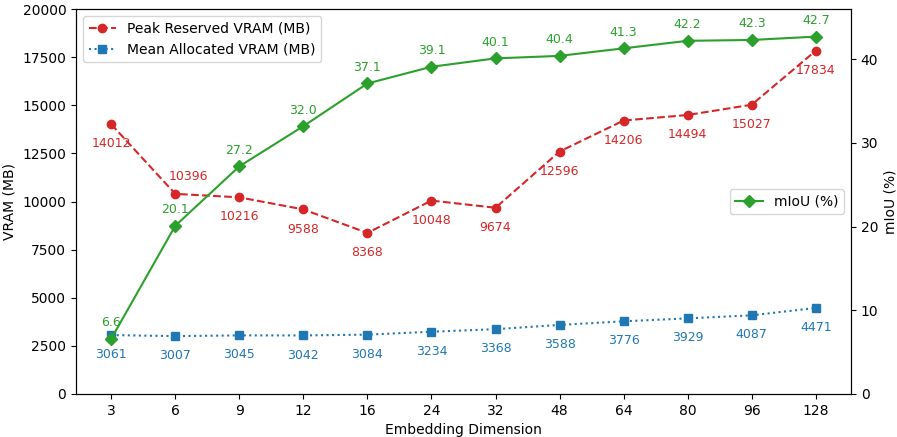}
    \caption{VRAM consumption across different feature compressor embedding dimensions}
    \label{fig:vram_plot}
\end{figure}

The results indicate that increasing the embedding dimensionality improves Cosine Similarity. However, after reaching 64 dimensions, these improvements become negligible, while processing time continues to rise. Additionally, larger embedding sizes introduce increased VRAM consumption, complexity in optimizing Gaussian parameters, leading to a decline in PSNR.

Interestingly, at an embedding dimensionality of 3, compression is highly lossy, resulting in poor reconstruction of feature embeddings. This leads to ineffective optimization of Gaussian parameters, causing an overgeneration of Gaussians as the pipeline attempts to compensate for lost information. This behavior significantly increases VRAM consumption (as shown in Figure~\ref{fig:vram_plot}) without yielding meaningful improvements in scene reconstruction or segmentation quality.

The selected dimensionality of 64 provides a favorable balance, preserving essential semantic information while maintaining efficient optimization and reconstruction quality. It ensures that feature embeddings retain sufficient expressiveness without unnecessary computational overhead.

\label{sec:experiments_comparison}
\section{Conclusion}
\label{sec:conclusion}

We introduced LEG-SLAM, a novel approach that combines 3D Gaussian Splatting with open-vocabulary semantic understanding, enabling real-time 3D reconstruction and interactive scene interpretation based on textual queries. Unlike existing methods, our approach does not rely on predefined object categories and provides efficient integration of semantic and geometric information.

Experiments demonstrate that LEG-SLAM achieves competitive reconstruction quality on the Replica dataset while maintaining high-speed performance for semantic segmentation on ScanNet. The use of DINOv2 for feature extraction and PCA for embedding compression ensures computational efficiency without significant loss of accuracy. However, integrating semantic information during optimization slightly affects reconstruction quality due to additional constraints in Gaussian optimization.

Furthermore, LEG-SLAM significantly accelerates open-vocabulary Gaussian Splatting-based segmentation. Compared to LERF, our method achieves a 30× speedup, making real-time interactive scene understanding feasible for practical applications.

Despite minor trade-offs in reconstruction quality due to semantic constraints, LEG-SLAM offers a unique balance between speed, adaptability, and real-time capability, making it a promising solution for future 3D scene understanding tasks.
{
    \small
    \bibliographystyle{ieeenat_fullname}
    \bibliography{main}
}

\clearpage
\setcounter{page}{1}
\setcounter{section}{0}
\setcounter{table}{0}
\setcounter{figure}{0}

\maketitlesupplementary

\section{ScanNet dataset}

To further analyze the performance of LEG-SLAM, we evaluate the reconstruction and semantic segmentation metrics on individual ScanNet scenes. Table~\ref{tab:supp_scannet_metrics} presents per-scene results, where PSNR, SSIM, and LPIPS measure reconstruction quality, while mIoU and mAcc assess semantic segmentation accuracy.

For semantic segmentation, LEG-SLAM operates in an open-vocabulary setting, but to quantitatively evaluate performance, we align with the standard ScanNet benchmark. The evaluation uses 20 semantic categories, which are detailed in Table~\ref{tab:supp_scannet_classes}, alongside their per-class IoU scores. 

To ensure effective feature compression, the same 20 categories were used to train the PCA encoder for ScanNet dataset evaluation. This alignment allows PCA to learn the optimal projection for retaining key semantic information while reducing computational complexity. However, since PCA learns a generalized compression strategy rather than memorizing specific class distributions, our method remains effective even on previously unseen categories. While segmentation quality is more consistent on trained categories, LEG-SLAM successfully extends to open-vocabulary scenarios, making it applicable across diverse 3D environments.

\begin{table}[!h]
\centering
\small
\resizebox{\columnwidth}{!}{%
\begin{tabular}{l|c|c|c|c|c}
\hline
\textbf{Scene} & \textbf{PSNR (dB)} & \textbf{SSIM} & \textbf{LPIPS} & \textbf{mIoU} & \textbf{mAcc} \\
\hline
0050\_02 & 21.51 & 0.6963 & 0.3399 & 0.3157 & 0.5474 \\
0144\_01 & 25.23 & 0.8167 & 0.1914 & 0.5388 & 0.6758 \\
0221\_01 & 23.69 & 0.7314 & 0.3868 & 0.4939 & 0.6852 \\
0300\_01 & 26.66 & 0.8410 & 0.2041 & 0.5340 & 0.6204 \\
0354\_00 & 20.98 & 0.8307 & 0.2048 & 0.5554 & 0.7378 \\
0389\_00 & 26.51 & 0.8302 & 0.2183 & 0.5382 & 0.8238 \\
0423\_02 & 23.14 & 0.7491 & 0.2386 & 0.4499 & 0.4633 \\
0427\_00 & 25.41 & 0.8080 & 0.2163 & 0.4985 & 0.6413 \\
0494\_00 & 24.96 & 0.8264 & 0.1902 & 0.5872 & 0.7972 \\
0616\_00 & 19.03 & 0.7386 & 0.3031 & 0.5872 & 0.8007 \\
0645\_02 & 22.03 & 0.7797 & 0.2892 & 0.4229 & 0.7224 \\
0693\_00 & 27.90 & 0.8394 & 0.1870 & 0.4115 & 0.7032 \\
\hline
\end{tabular}%
}
\caption{Quantitative evaluation on ScanNet scenes}
\label{tab:supp_scannet_metrics}
\end{table}

\begin{table}[ht]
\centering
\small
\begin{tabular}{l|c}
\hline
\textbf{Class} & \textbf{IoU} \\
\hline
Wall          & 0.364  \\
Floor         & 0.729  \\
Cabinet       & 0.198  \\
Bed           & 0.759  \\
Chair         & 0.497  \\
Sofa          & 0.275  \\
Table         & 0.577  \\
Door          & 0.450  \\
Window        & 0.441  \\
Shelves       & 0.521  \\
Counter       & not presented  \\
Curtain       & 0.626  \\
Ceiling       & 0.338  \\
Refrigerator  & 0.385  \\
Television    & 0.394  \\
Person        & not presented \\
Toilet        & 0.471  \\
Sink          & 0.034  \\
Lamp          & 0.219  \\
Bag           & 0.181  \\
\hline
\textbf{Mean IoU} & \textbf{0.414} \\
\textbf{Mean Accuracy} & \textbf{0.743} \\
\hline
\end{tabular}%
\caption{Per-class Intersection over Union (IoU) scores for ScanNet dataset. Some classes (\textit{counter}, \textit{person}) were not evaluated due to annotation errors.}
\label{tab:supp_scannet_classes}
\end{table}
\section{Replica dataset}

To contextualize LEG-SLAM’s performance, we compare it against existing SLAM methods on different Replica scenes. Table~\ref{tab:replica_scenes_comparison} summarizes the results, providing a detailed breakdown of reconstruction quality across multiple environments. The table reports PSNR, SSIM, and LPIPS metrics, averaged over all scenes and shown individually for each tested environment.

\begin{table*}[ht!]
\centering
\scriptsize
\resizebox{0.85\textwidth}{!}{%
\begin{tabular}{l|l|c|c|c|c|c|c|c|c|c}
\hline
\textbf{Methods} & \textbf{Metrics} & \textbf{Avg.} & \textbf{Room0} & \textbf{Room1} & \textbf{Room2} & \textbf{Office0} & \textbf{Office1} & \textbf{Office2} & \textbf{Office3} & \textbf{Office4} \\ \hline
\multirow{3}{*}{NICE-SLAM} 
& PSNR$\uparrow$ & 24.42 & 22.12 & 22.47 & 24.52 & 29.07 & 30.34 & 19.66 & 22.23 & 24.94 \\
& SSIM$\uparrow$ & 0.809 & 0.689 & 0.757 & 0.814 & 0.874 & 0.886 & 0.797 & 0.801 & 0.856 \\
& LPIPS$\downarrow$ & 0.233 & 0.330 & 0.271 & 0.208 & 0.229 & 0.181 & 0.235 & 0.209 & 0.198 \\ \hline

\multirow{3}{*}{Co-SLAM} 
& PSNR$\uparrow$ & 30.24 & 27.27 & 28.45 & 29.06 & 34.14 & 34.87 & 28.43 & 28.76 & 30.91 \\
& SSIM$\uparrow$ & 0.939 & 0.910 & 0.909 & 0.932 & 0.961 & 0.969 & 0.938 & 0.941 & 0.955 \\
& LPIPS$\downarrow$ & 0.252 & 0.324 & 0.294 & 0.266 & 0.209 & 0.196 & 0.258 & 0.229 & 0.236 \\ \hline

\multirow{3}{*}{ESLAM} 
& PSNR$\uparrow$ & 29.08 & 25.32 & 27.77 & 29.08 & 33.71 & 30.20 & 28.09 & 28.77 & 29.71 \\
& SSIM$\uparrow$ & 0.929 & 0.875 & 0.902 & 0.932 & 0.960 & 0.923 & 0.943 & 0.948 & 0.945 \\
& LPIPS$\downarrow$ & 0.336 & 0.313 & 0.298 & 0.248 & 0.184 & 0.228 & 0.241 & 0.196 & 0.204 \\ \hline

\multirow{3}{*}{SplaTAM} 
& PSNR$\uparrow$ & 33.98 & 32.48 & 33.72 & 34.96 & 38.34 & 39.04 & 31.90 & 29.70 & 31.68 \\
& SSIM$\uparrow$ & \textbf{0.969} & 0.975 & 0.970 & \textbf{0.982} & 0.982 & \textbf{0.982} & 0.965 & 0.950 & 0.946 \\
& LPIPS$\downarrow$ & \textbf{0.099} & 0.072 & 0.096 & 0.074 & \textbf{0.083} & 0.093 & \textbf{0.100} & 0.118 & 0.155 \\ \hline

\multirow{3}{*}{SGS-SLAM} 
& PSNR$\uparrow$ & \textbf{34.66} & 32.50 & 34.25 & \textbf{35.10} & \textbf{38.54} & \textbf{39.20} & \textbf{32.90} & \textbf{32.05} & \textbf{32.75} \\
& SSIM$\uparrow$ & \textbf{0.973} & \textbf{0.976} & \textbf{0.978} & 0.981 & \textbf{0.984} & 0.980 & 0.967 & 0.966 & 0.949 \\
& LPIPS$\downarrow$ & 0.096 & \textbf{0.070} & 0.094 & \textbf{0.070} & 0.086 & \textbf{0.087} & 0.101 & \textbf{0.115} & \textbf{0.148} \\ \hline

\multirow{3}{*}{Ours} 
& PSNR$\uparrow$ & 32.94 & 27.92 & 30.89 & 32.77 & 36.19 & 36.99 & 29.23 & 31.47 & 33.57 \\
& SSIM$\uparrow$ & 0.914 & 0.8246 & 0.9028 & 0.9312 & 0.9509 & 0.9409 & 0.9229 & 0.9208 & 0.9367 \\
& LPIPS$\downarrow$ & 0.148 & 0.2081 & 0.1409 & 0.1190 & 0.1310 & 0.1439 & 0.1685 & 0.1460 & 0.1269 \\ \hline

\end{tabular}
}

\caption{Comparison of reconstruction quality of SLAM methods on different Replica scenes.}

\label{tab:replica_scenes_comparison}
\end{table*}
\begin{figure*}[htbp]
    \centering
    \resizebox{0.75\textwidth}{!}{%
        \begin{minipage}{\textwidth}
            \begin{subfigure}{0.32\textwidth}
                \includegraphics[width=\linewidth]{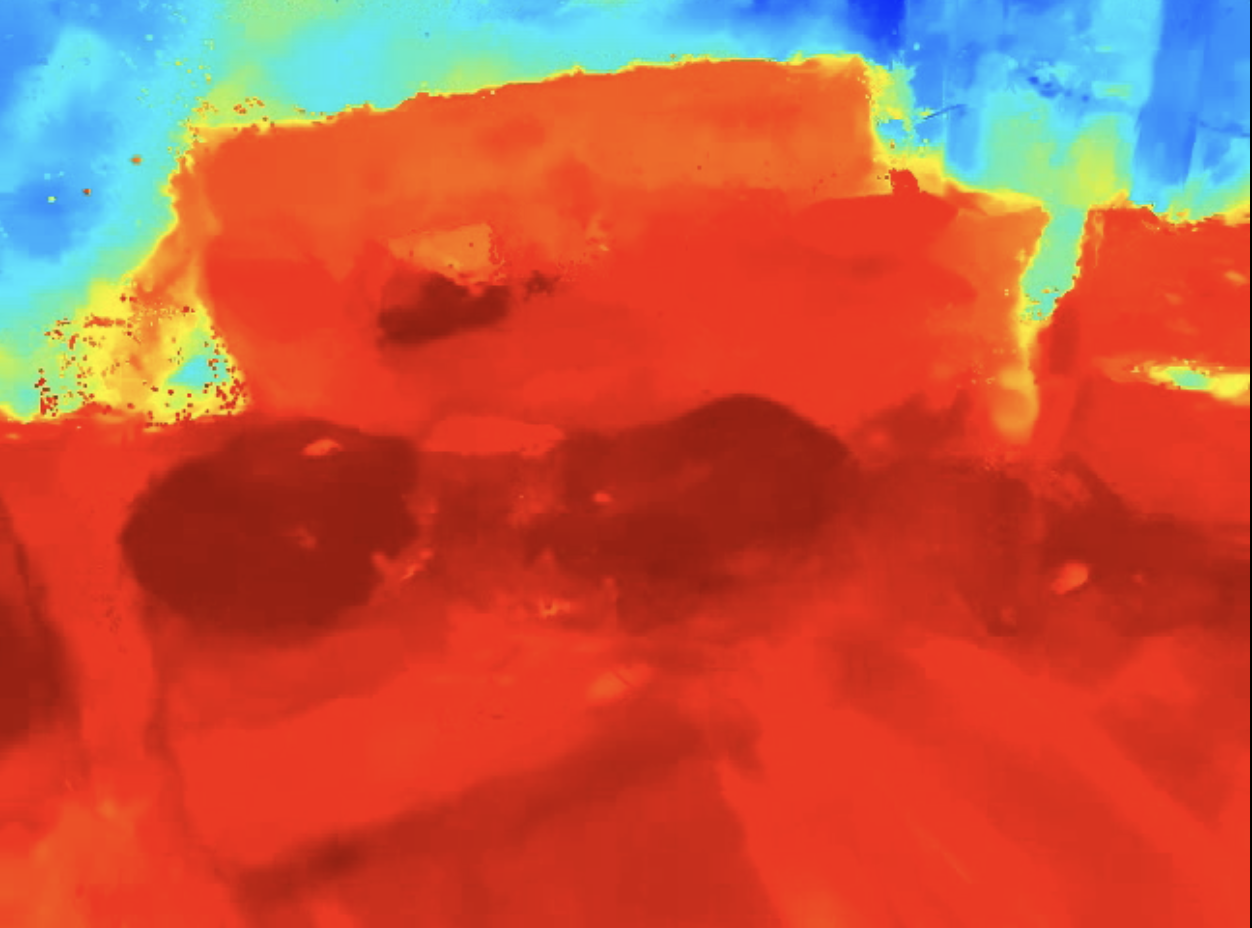}
                \caption{Dim = 3}
            \end{subfigure}
            \begin{subfigure}{0.32\textwidth}
                \includegraphics[width=\linewidth]{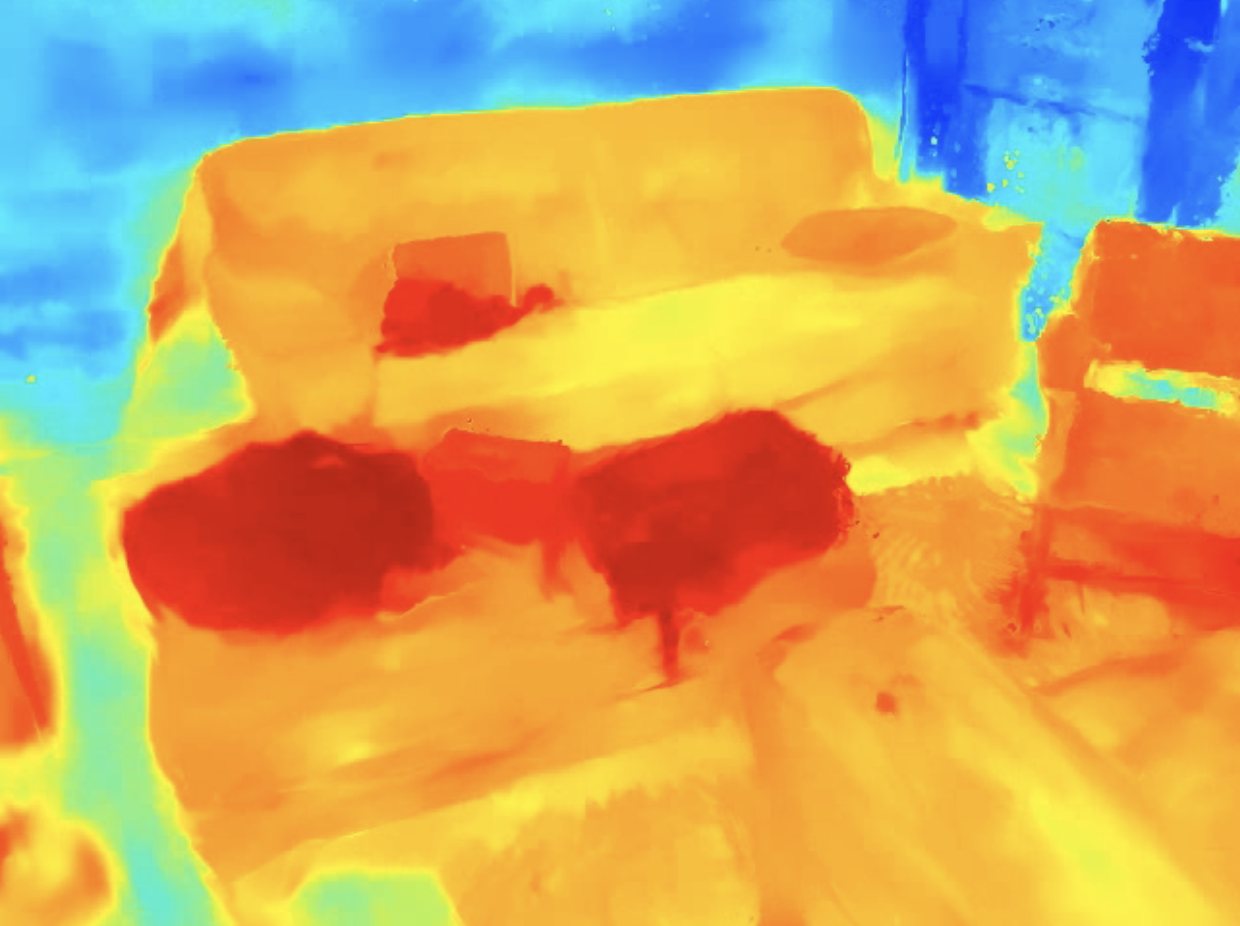}
                \caption{Dim = 6}
            \end{subfigure}
            \begin{subfigure}{0.32\textwidth}
                \includegraphics[width=\linewidth]{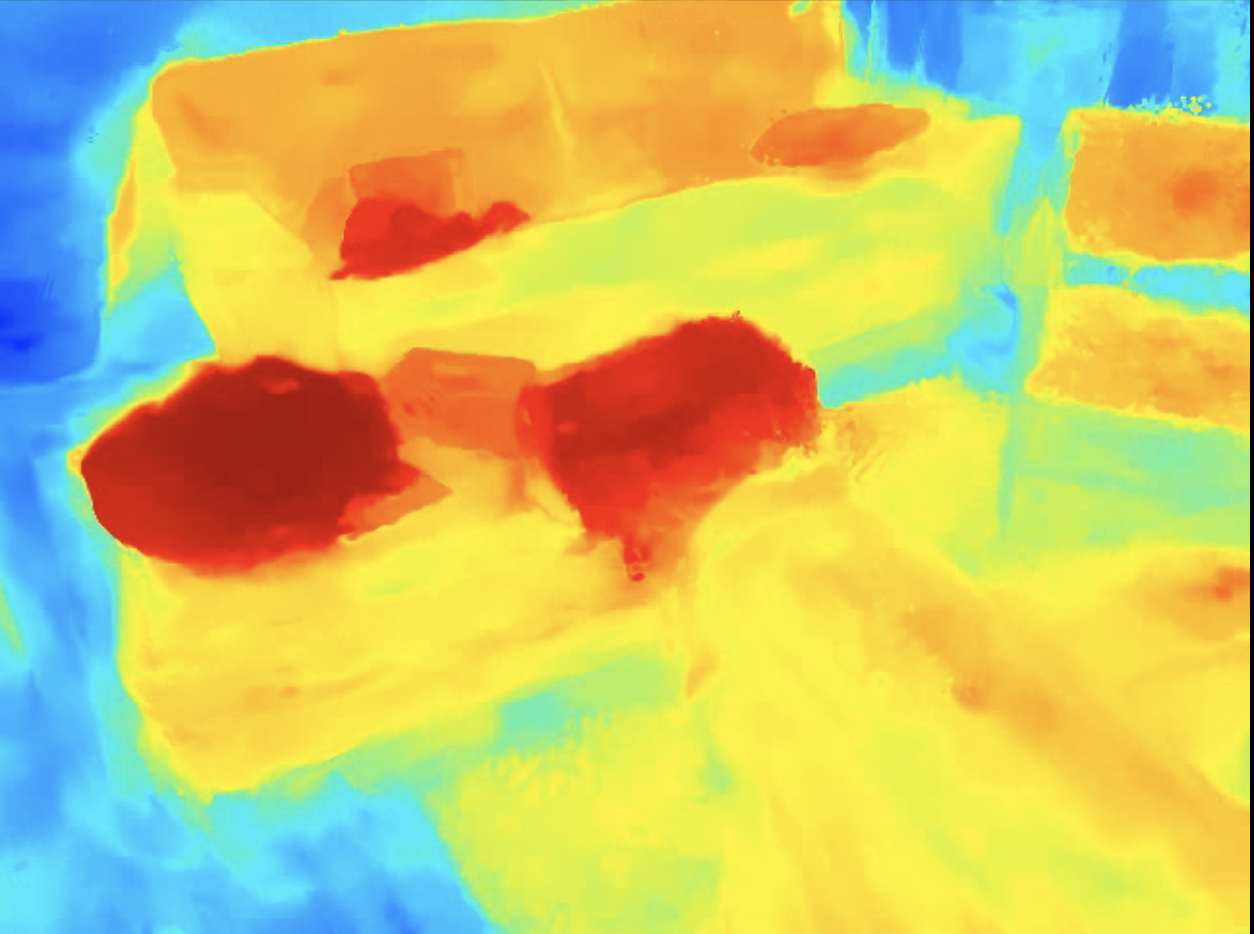}
                \caption{Dim = 9}
            \end{subfigure}

            \begin{subfigure}{0.32\textwidth}
                \includegraphics[width=\linewidth]{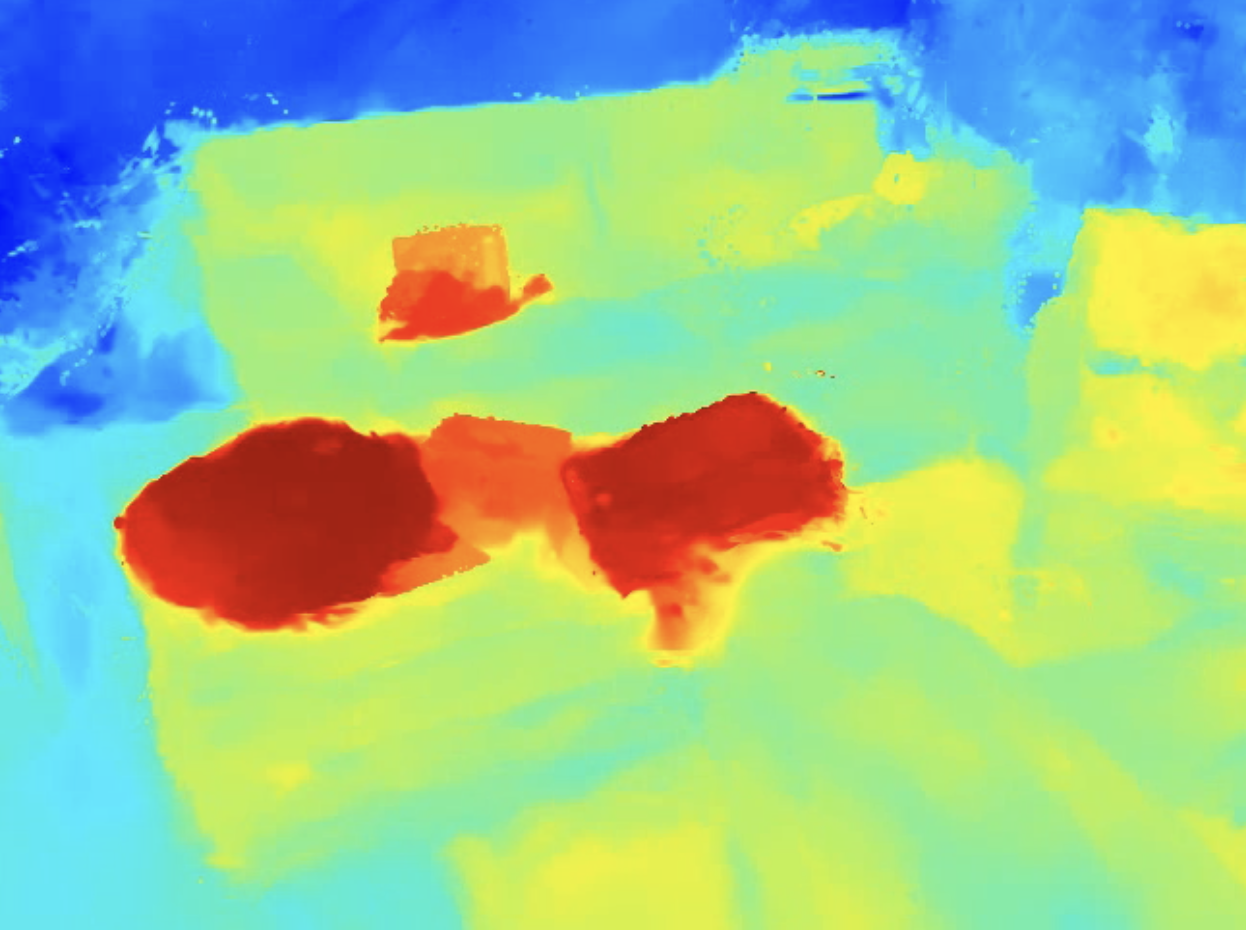}
                \caption{Dim = 12}
            \end{subfigure}
            \begin{subfigure}{0.32\textwidth}
                \includegraphics[width=\linewidth]{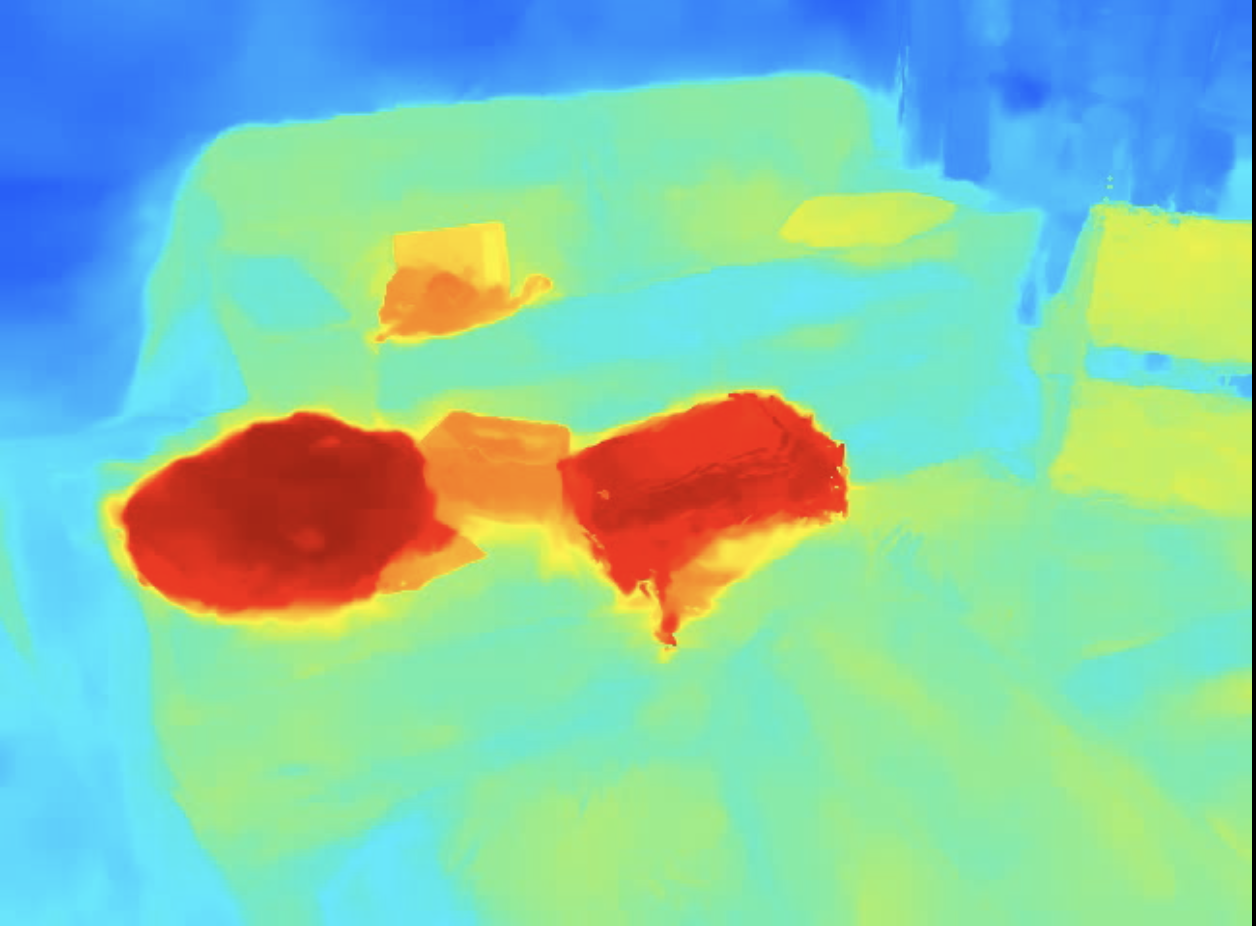}
                \caption{Dim = 16}
            \end{subfigure}
            \begin{subfigure}{0.32\textwidth}
                \includegraphics[width=\linewidth]{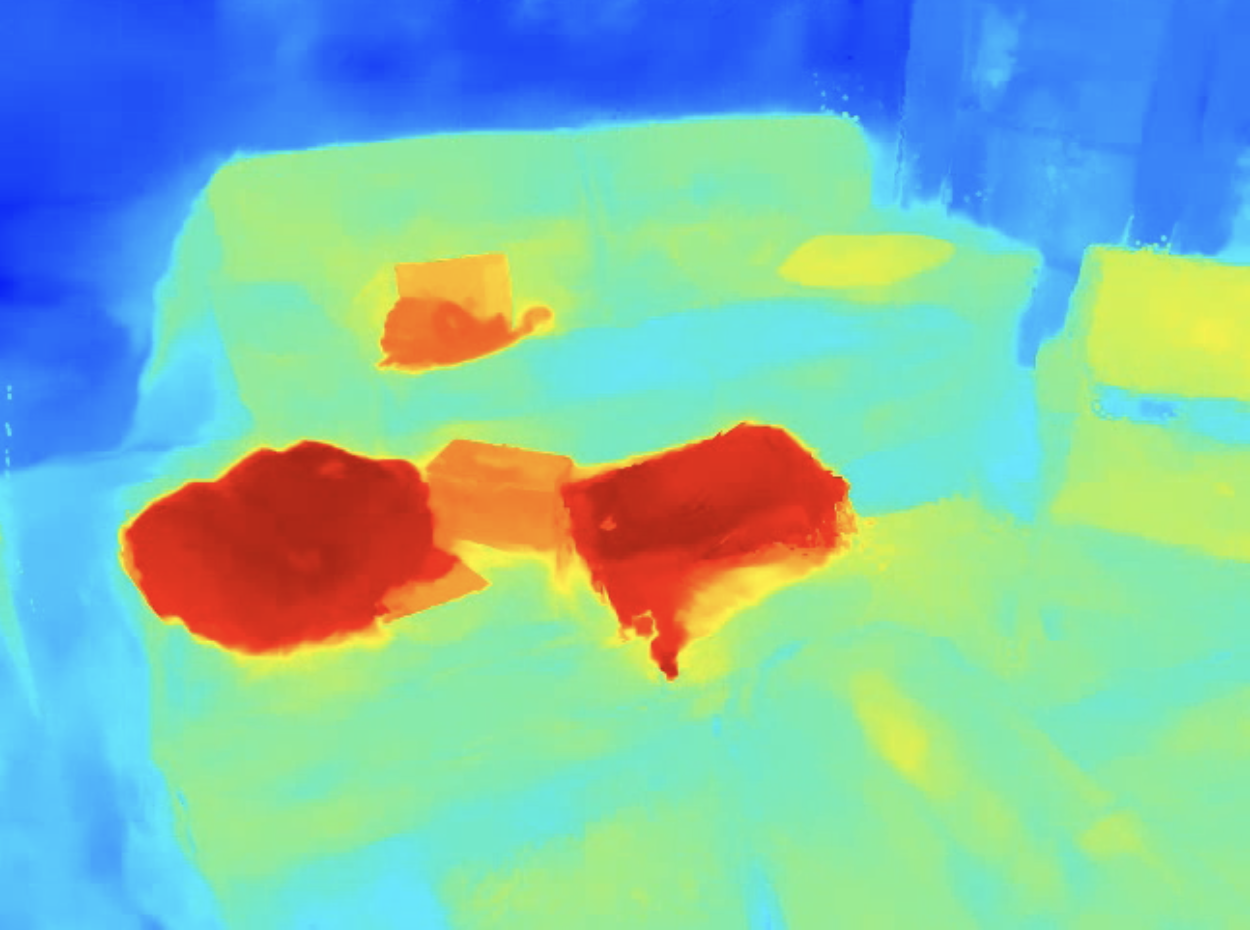}
                \caption{Dim = 24}
            \end{subfigure}

            \begin{subfigure}{0.32\textwidth}
                \includegraphics[width=\linewidth]{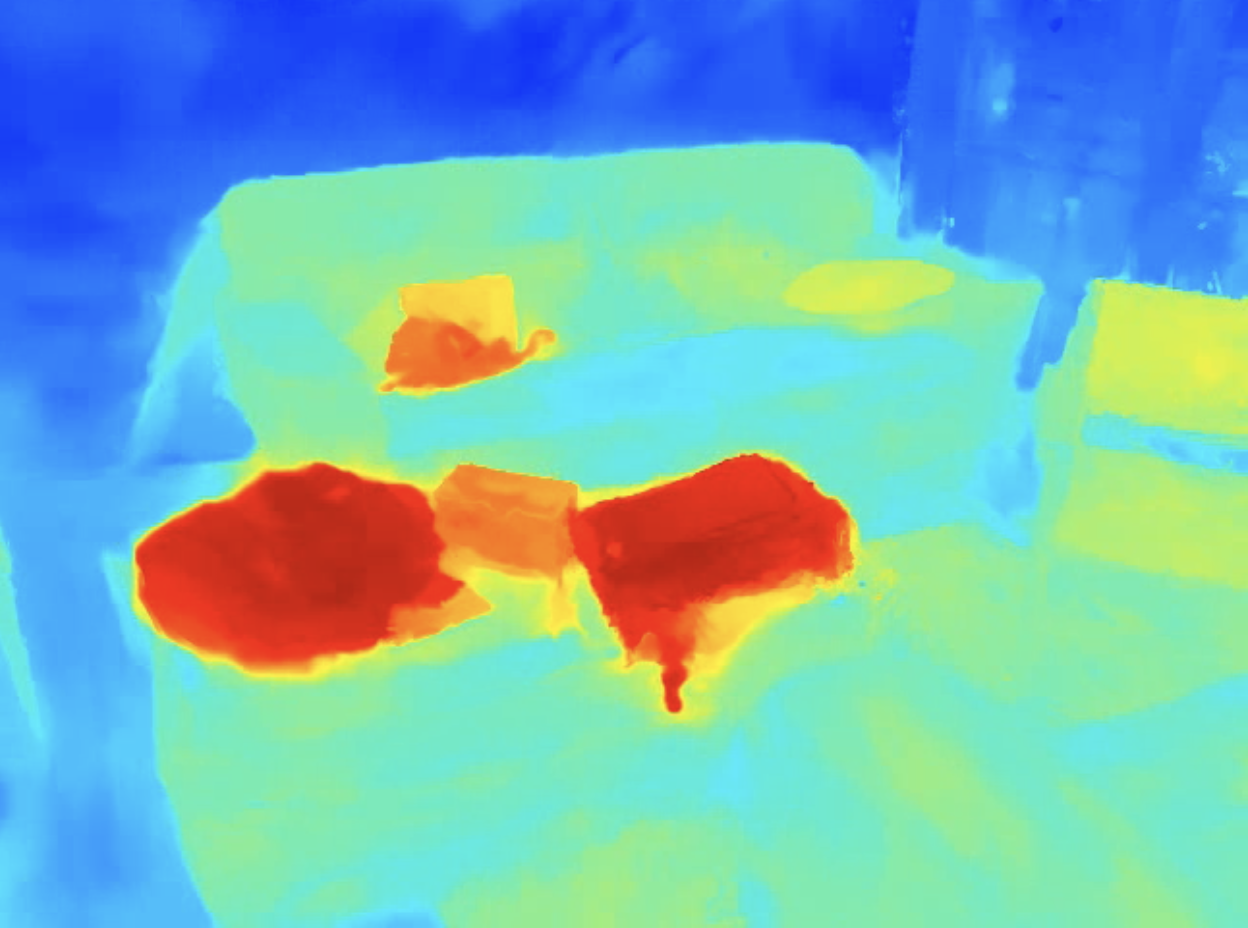}
                \caption{Dim = 32}
            \end{subfigure}
            \begin{subfigure}{0.32\textwidth}
                \includegraphics[width=\linewidth]{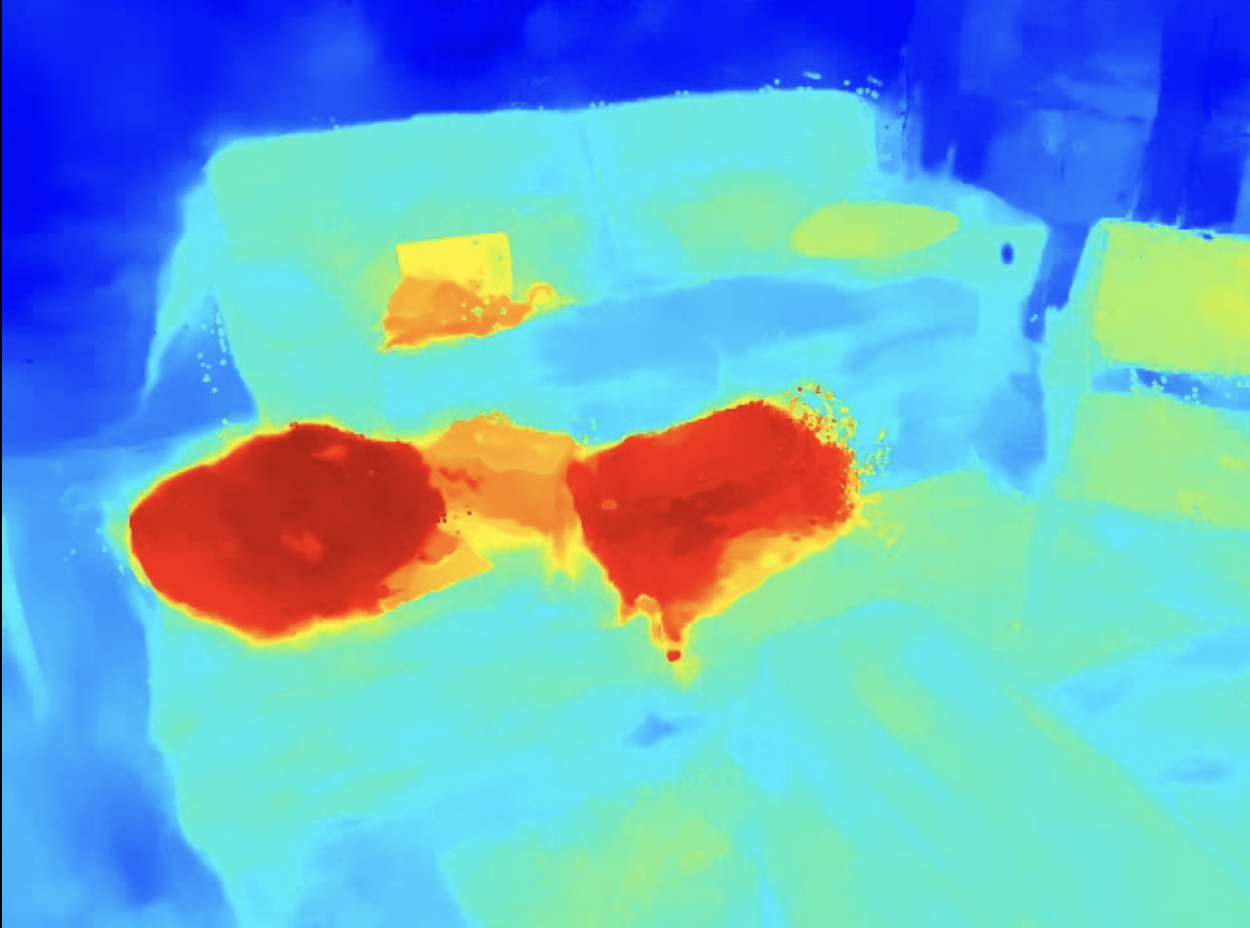}
                \caption{Dim = 48}
            \end{subfigure}
            \begin{subfigure}{0.32\textwidth}
                \includegraphics[width=\linewidth]{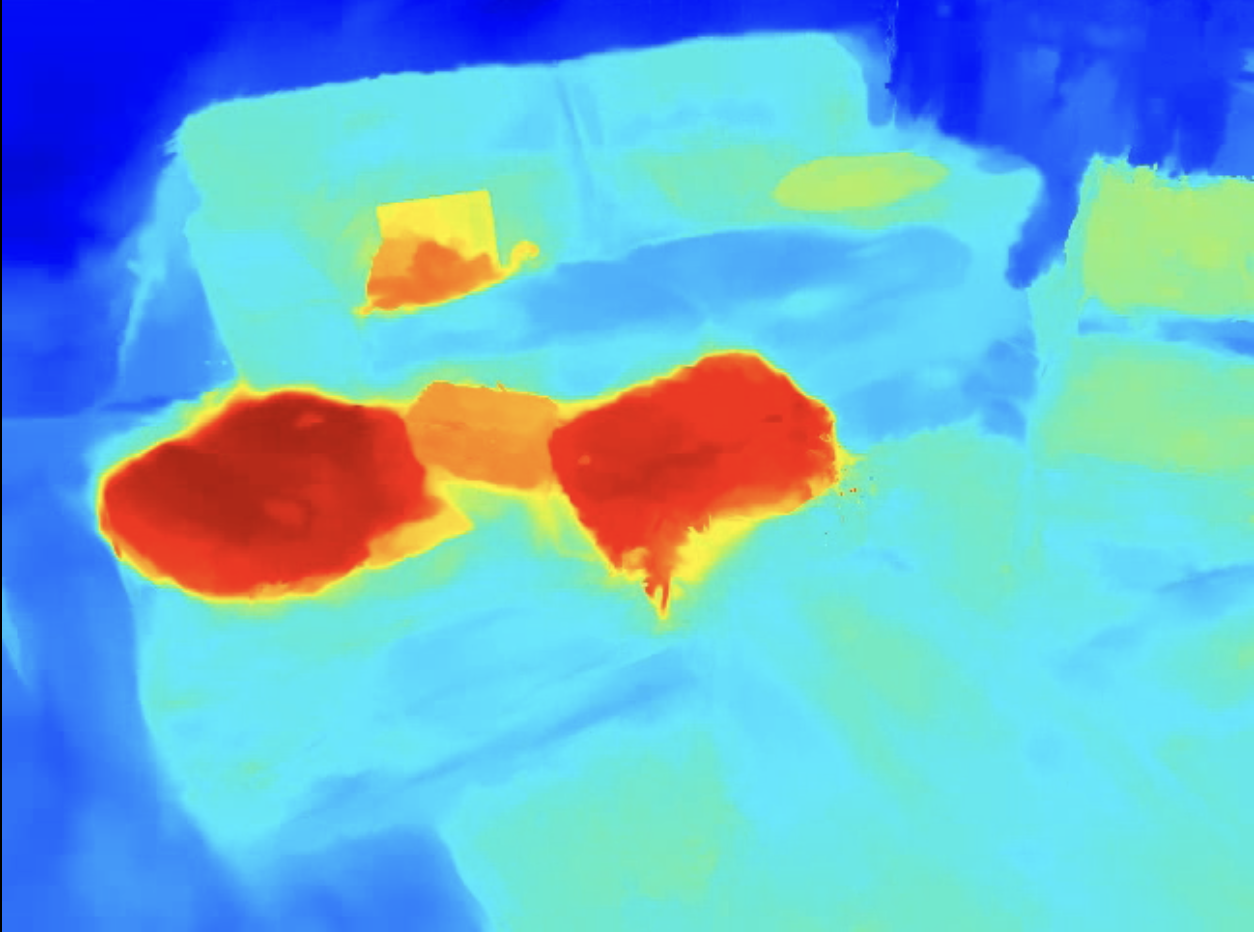}
                \caption{Dim = 64}
            \end{subfigure}
        \end{minipage}
    }
    \caption{Comparison of PCA embedding compression on the \textbf{backpack} query across different dimensions. Each row represents different compression levels (3–64), demonstrating how embedding dimensionality affects reconstruction quality and semantic consistency.}
    \label{fig:supplementary_backpach}
\end{figure*}

\section{Impact of Embedding Dimensionality on Semantic Quality}

To analyze the effect of embedding compression, we compare the semantic reconstruction quality across different PCA embedding dimensions, as shown in Figure~\ref{fig:supplementary_backpach}. The visualizations illustrate how varying the dimensionality from 3 to 64 affects the ability to preserve fine-grained semantic details while maintaining computational efficiency.

Lower-dimensional embeddings (3–9 dimensions) result in a significant loss of semantic information, leading to blurry segmentations and imprecise object boundaries. The compression at this level fails to retain fine details, causing objects to merge and reducing the distinctiveness of different semantic regions. As the dimensionality increases to 12–24, the quality of segmentation improves considerably. The reconstructed semantic maps become more structured, with clearer object outlines and more accurate spatial distributions. However, some artifacts remain, particularly in complex scene regions where fine-grained features are necessary for precise segmentation.

At higher embedding dimensions (32–64), the semantic reconstructions closely resemble the original feature space. Objects maintain well-defined contours, and the color-coded segmentations demonstrate a high level of semantic accuracy. The 64-dimensional case provides the most faithful reconstruction, preserving both large-scale structure and fine object details. However, beyond this point, the benefits of increasing dimensionality further become negligible while computational overhead continues to grow. 

These results confirm that PCA effectively retains essential semantic information while ensuring computational efficiency. The selection of 64 as the embedding dimension achieves a balance between compression quality, segmentation accuracy, and real-time performance, making it the optimal choice for LEG-SLAM.

\end{document}